\documentclass[10pt,twocolumn,letterpaper]{article}

\usepackage{ijcb}
\usepackage{times}
\usepackage{epsfig}
\usepackage{graphicx}
\usepackage{amsmath}
\usepackage{amssymb}
\usepackage{caption}
\usepackage{algorithm}
\usepackage{algpseudocode}
\usepackage{subcaption}
\usepackage[dvipsnames]{xcolor}

\usepackage[pagebackref=true,breaklinks=true,letterpaper=true,colorlinks,bookmarks=false]{hyperref}

\ijcbfinalcopy % *** Uncomment this line for the final submission

\begin{document}

%%%%%%%%% TITLE
%\title{On generating responsible and fair StyleGAN-based complementary dataset for face-recognition systems}
% Other ideas for the title ...
%\title{Tackling Fairness with Synthetic Dataset Generation}
%\title{A step toward responsible and fair dataset using controlled synthetic image generation} 

%\title{Bridging the Fairness Gap:\\StyleGAN-based Complementary Synthesized Datasets for Face-Recognition}
% I think this title is more attractive and also similar to what you said. also we refer to StyleGAN

\title{Toward responsible face datasets: modeling the distribution of a disentangled latent space for sampling face images from demographic groups}

\author{Parsa Rahimi$^{1,2}$, Christophe Ecabert$^{1}$, Sébastien Marcel$^{1,3}$\\
\\
$^{1}$Idiap Research Institute, $^{2}$École Polytechnique Fédérale de Lausanne (EPFL),\\$^{3}$School of Criminal Justice, Université de Lausanne (UNIL)\\
{\tt\small parsa.rahiminoshanagh@epfl.ch, christophe.ecabert@idiap.ch, sebastien.marcel@idiap.ch}\\
% \and Christophe Ecabert\\
% \\
% Idiap Research Institute \\
% {\tt\small christophe.ecabert@idiap.ch}
% \and
% Sébastien Marcel\\
% \\
% Idiap Research Institute, UNIL \\
% {\tt\small sebastien.marcel@idiap.ch}
}

\maketitle
\thispagestyle{empty}

%%%%%%%%% ABSTRACT
\begin{abstract}
    Recently, it has been exposed that some modern facial recognition systems could discriminate specific demographic groups and may lead to unfair attention with respect to various facial attributes such as gender and origin. The main reason are the biases inside datasets, unbalanced demographics, used to train theses models. Unfortunately, collecting a large-scale balanced dataset with respect to various demographics is impracticable.
    In this paper, we investigate as an alternative the generation of a balanced and possibly bias-free synthetic dataset that could be used to train, to regularize or to evaluate deep learning-based facial recognition models. We propose to use a simple method for modeling and sampling a disentangled projection of a StyleGAN latent space to generate any combination of  demographic groups (e.g. $hispanic-female$). Our experiments show that we can synthesis any combination of demographic groups effectively and the identities are different from the original training dataset. We also released the source code \footnote{\url{https://gitlab.idiap.ch/biometric/sg_latent_modeling}}.
    % In this paper, we are trying to address these biases with complementary fair synthetic dataset generation for the fair face recognition problem. 
    
    %\sug{Make the argument more general? not only face? } 
\end{abstract}

%%%%%%%%% BODY TEXT
\section{Introduction}
\label{sec:intro}

\begin{figure}[!htb]
   \centering
   \begin{subfigure}[b]{0.49\linewidth}
       \centering
       \includegraphics[width=\linewidth]{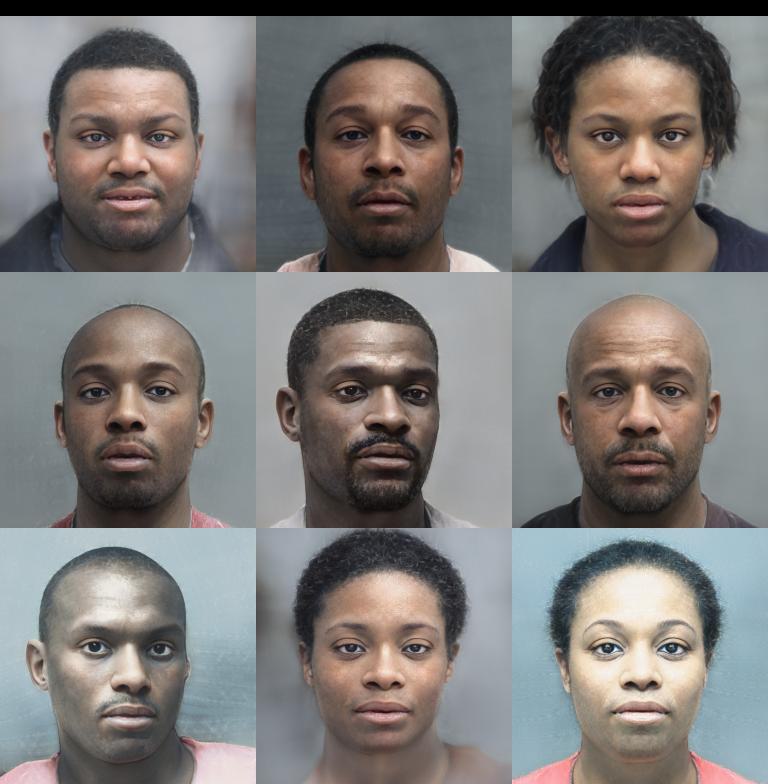}
       \caption{\emph{African-American}}
   \end{subfigure}
   \begin{subfigure}[b]{0.49\linewidth}
       \centering
       \includegraphics[width=\linewidth]{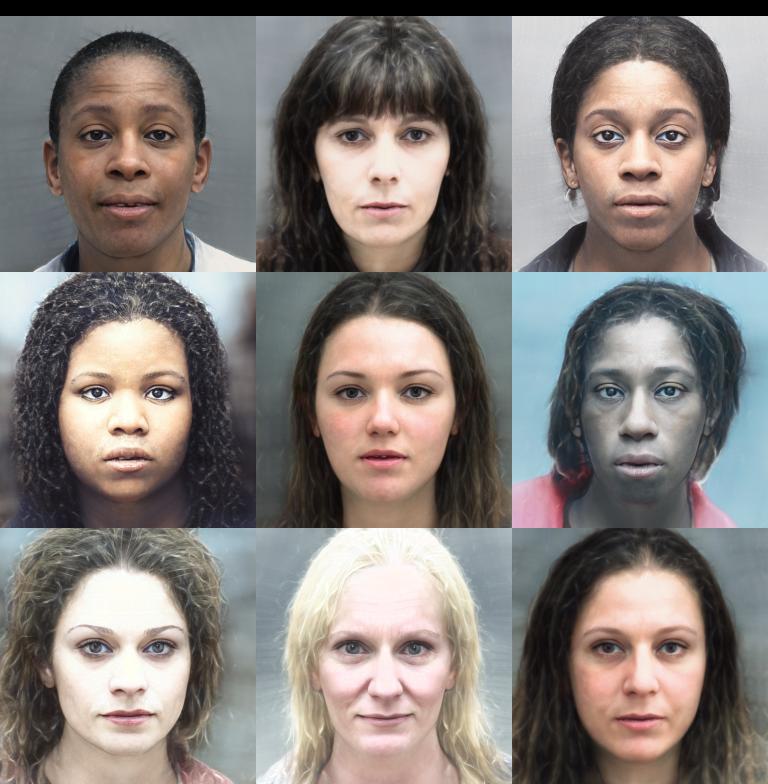}
       \caption{\emph{Female}} 
   \end{subfigure}
   \begin{subfigure}[b]{0.49\linewidth}
       \centering
       \includegraphics[width=\linewidth]{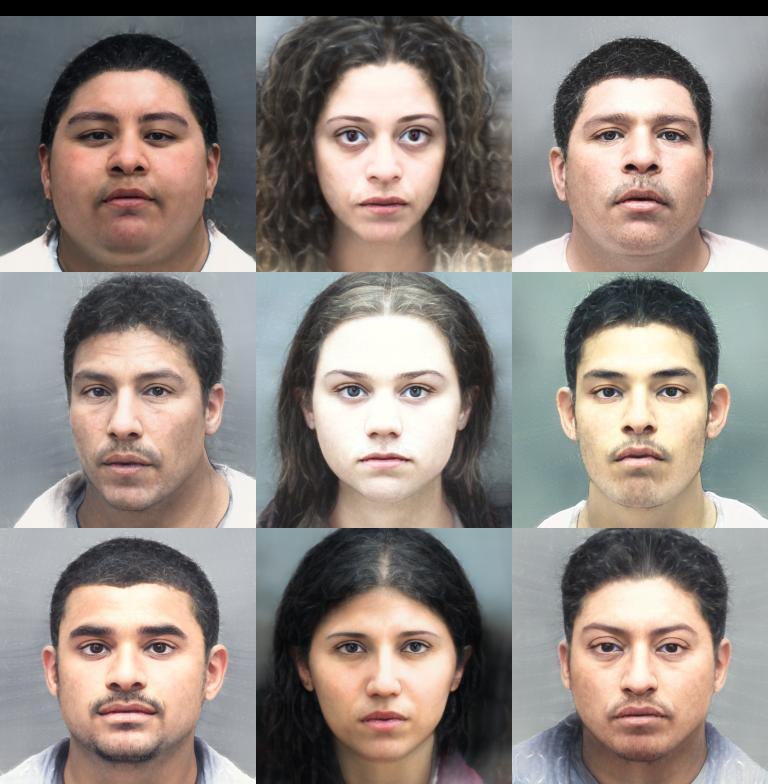}
       \caption{\emph{Hispanic}}
   \end{subfigure}
   \begin{subfigure}[b]{0.49\linewidth}
       \centering
       \includegraphics[width=\linewidth]{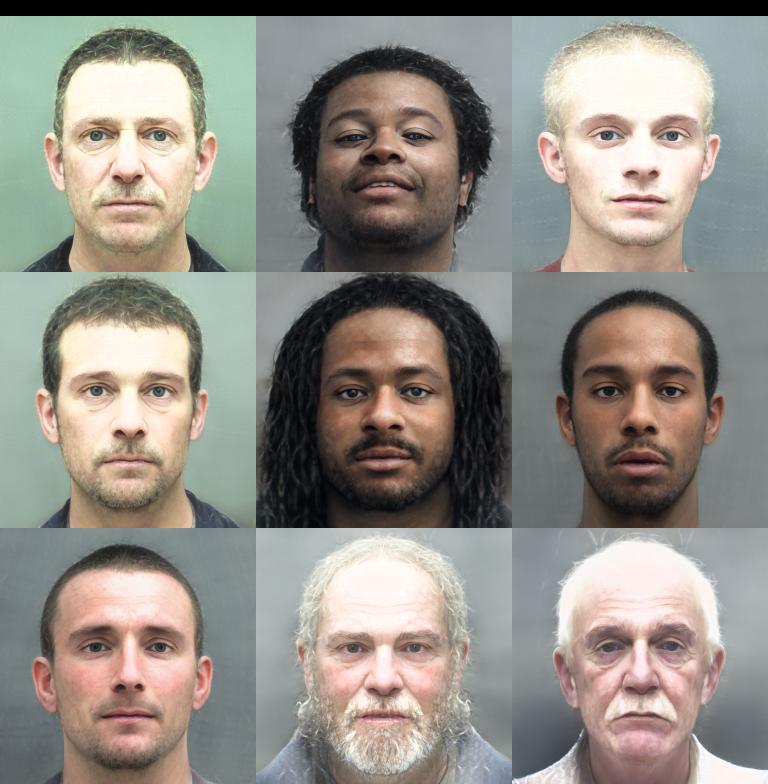}
        \caption{\emph{Male}}
   \end{subfigure}
   \caption{Generated face images according to desired demographic groups, each 3x3 tile shows images sampled from different demographic groups.}
    \label{fig:on_demand_generated_demogrpahics}
    \vspace{-15pt}
\end{figure}

\begin{figure*}
  \centering
  \includegraphics[width=1.0\textwidth]{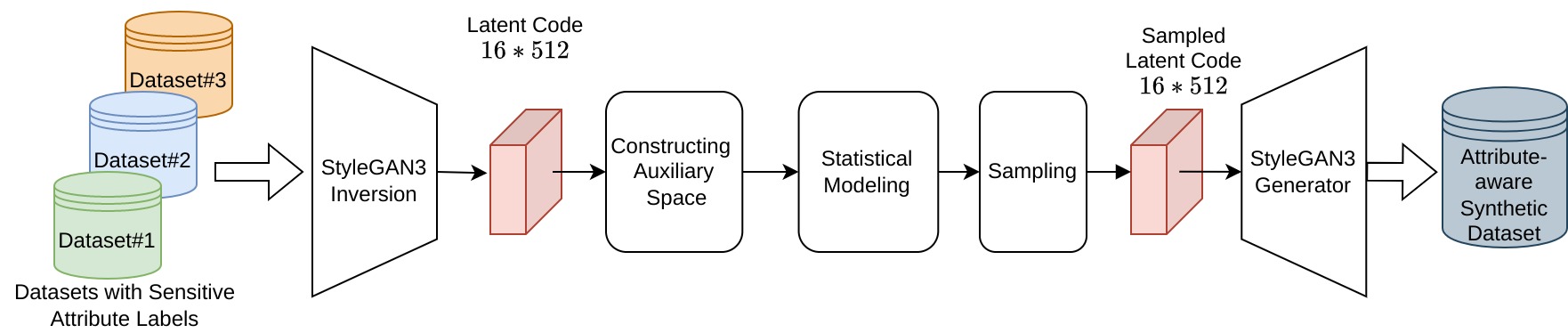}
  \caption{Overall pipeline of our proposed method. Starting from datasets with demogrpahic labels. Usin StyleGAN inversion, we invert the images to desired latent space. To facilitate modeling of StyleGAN latent space we build disentangled auxilarity space. We sample the modeled space to generate desired demogrphic.}
  \label{figure:overall_pipeline} 
  \vspace{-10pt}
\end{figure*}   

The use of face recognition (FR) systems in critical applications such as law enforcement and recruitment has raised significant ethical concerns. Recent studies have demonstrated that commercially available FR systems using Artificial Intelligence (AI) can exhibit unfairness and bias, particularly against certain demographic groups \cite{buolamwini_gender_2018, rosenberg_fairness_2022}. As AI systems become more widely adopted in our daily lives, addressing these ethical and legal considerations becomes even more important.

In many cases, the use of FR technology is subject to legal restrictions and regulations, further highlighting the importance of developing fair and accurate systems. One of the main challenges in achieving fairness in FR is the lack of diverse training data, especially considering that the key reason for the success of recent large FR networks is the large datasets which they are being trained upon. At the same time, due to legal and ethical grounds, most of the widely used FR datasets like MS-Celeb1M \cite{msceleb}, VGGFace2 \cite{cao2018vggface2} and MegaFace \cite{kemelmacher2016megaface} 
have been retracted. Also considering legal policies such as \cite{eugdrppolicy, dicagealg}, usage of existing datasets including WebFace260M \cite{zhu2021webface260m} and CASIA-WebFace \cite{casiawebface} might also become troublesome when they are deployed in critical applications. Besides these concerns, collecting large amounts of samples required to train deep FR models with various  balanced demographic groups is another problem. Therefore, developing complementary datasets that accurately represent underrepresented groups is crucial in mitigating these issues. 

% with the aim of reducing bias and improving the fairness and accuracy of facial recognition systems.

The purpose of this work is the creation of balanced face datasets to reduce the bias of the FR models. Currently, approaches for bias mitigation in FR include \textit{pre-processing}, \textit{in-processing}, and \textit{post-processing}. Pre-processing approaches involve modifying the input data to remove or reduce the effects of bias \cite{kamiran2012data, calmon2017optimized, bellamy2019ai}. In-processing approaches work by changing the model architecture or learning algorithm to make it more robust to bias \cite{louizos2015variational, louppe2017learning, wadsworth2018achieving, zafar2017fairness, zhang2018mitigating, celis2019classification, razeghi2022bottlenecks}. However, this can compromise fairness, model performance and can be computationally expensive. Post-processing approaches involve adjusting model predictions after training to make them more fair \cite{kamiran2010discrimination, hardt2016equality,pleiss2017fairness, chen2019fairness, alghamdi2022beyond}, but it can also create a trade-off between fairness and model performance and can be limited by model interpretability.

% Recent studies have proposed the use of synthetic data to address the fairness and accuracy challenges in face recognition systems \cite{controlable_3dface_synthesis, cgofpp}. 

To address the lack of diversity in existing FR datasets, recent studies propose the use of synthetic data to reduce bias and improve accuracy \cite{controlable_3dface_synthesis, cgofpp}.
However, many of these approaches rely on randomly sampling the latent space of generator models and later attempting to steer and edit the generated signal to meet desired demographics~\cite{laurent}. This can result in accumulation of errors and further biases as the generation is not initially aware of demographic groups. To overcome this limitation and to address the lack of diversity in existing FR datasets, in this paper, we propose a novel yet simple approach to generate such a complementary dataset for FR systems. \autoref{fig:on_demand_generated_demogrpahics} show synthetic examples generated by our proposed method.
Our generation rely on StyleGAN-based ~\cite{sg3,sg2} models. This is mostly due to privacy concerns regarding diffusion-based generation. Indeed it was shown in \cite{extractingdiffusion23} that training data can be inferred from diffusion models which is a limitation for our application scenario to generate new face images.

% regarding extracting the training data for the diffusion models \cite{extractingdiffusion23}, which contradicts our privacy concerns in mind.

The proposed method can be expanded to any latent space-based generation architecture. One can see our method as the first step of any demographic editing methods. As we sample desired demographic groups equally, and later on we can employ editing methods like \cite{eg3d,controlable_3dface_synthesis,li2023cateccv22,laurent} to further generate different variations of same identity to introduce even larger fair datasets.

% Our approach leverages the frequency content of the generated images to produce more realistic and diverse synthetic data that accurately represents underrepresented groups. The goal of our approach is to bridge the fairness gap in face recognition by reducing biases in these systems. In this paper, we describe our methodology, experimental results, and discuss the implications of our findings for the field of face recognition.

%%%%%%%%%%%%%%%%%%%%%%%%%%%%%%%%%%%%%%%%%%%%%%%%%%%%%%%%
% MAPPING SECTION
%%%%%%%%%%%%%%%%%%%%%%%%%%%%%%%%%%%%%%%%%%%%%%%%%%%%%%%%
In \autoref{sec:related_works} we present related works in the domain of controlled generation and editing of face images. In \autoref{sec:proposed_method} we present our approach for controlled face generation. Finally in \autoref{sec:experiments} we validate our proposed method by various face-related tasks (e.g., demographic classification and identity experiment). 

% \sug{To talk about pre-prcessing, in-processinga and post-processing approaches }
% I added . 

% Comment these lines after checking 

% \textcolor{gray}{
% Following studies that demonstrate unfair commercial systems in critical applications
% \cite{buolamwini_gender_2018}w
% \cite{rosenberg_fairness_2022} % performance disparities with respect to demographics
% }
% \textcolor{gray}{
% And widely adopted AI systems in our daily life, addressing ethical and often law restrictive 
% }

% \textcolor{gray}{
% Most of the literature in recent years are seeing the problem synthesis as first randomly sampling the latent space of generator models e.g.  
% \cite{controlable_3dface_synthesis}
% \cite{cgofpp}
% and later on, try to steer and edit the generated signal to desired meet we the desired attribution, this might result in an accumulation error 
% %TODO cite something here
% and further biases as the generation in the first place is not aware of some sensitive groups.
% }

% \textcolor{gray}{
% \cite{sg3models_thirdtime}
% }

%%%%%%%%% Related works
\section{Related Works} \label{sec:related_works}

This section focuses on related works in controlled generation. Additionally, we provide a brief introduction to StyleGAN inversion methods in \autoref{sec:ganinversion}. After examining these methods, it becomes clear that not all of them are suitable for our particular needs.
% only one type is suitable for our particular needs.

\subsection{Prompt-based Synthesis Methods} 
Recent advances in generative models especially in diffusion based synthesis \cite{stablediff} and their ability to convert text to often realistic images brought new ways of exploration of generative models. As mentioned previously these methods are often pruned to privacy concerns and also exhibit uncontrollable output.
%One of the downsides of such models is their privacy concerns and oftentimes uncontrollable output

% i.e. our trials shows that even with some heavily engineered prompt the outcome might not correspond to what actually requested in the prompt. 

Using off-the-shelf models (e.g. FairFace classifier \cite{grover2019fair} and text-image encoders \cite{clip2021}),
\cite{genprompt22} modeled any control (text-based using CLIP, classifier-based using FairFace classifier ) via an energy-based model and try to minimize the divergence between the condition and the supervision of the auxiliary models. By introducing momentum constraint, authors in \cite{genprompt22} represented a debiased version of an arbitrary generator.

\subsection{Latent-Modeling Methods}
Authors in \cite{shukor_semantic_2022} suggested an autoencoder using normalizing flows \cite{realnvp} to form an auxiliary linear separable space. Later one can sample the new space and generate desired demographic groups.
%As they did not publish their code, we could not verify the results of this paper.

\cite{wei_latent_2022} first randomly sampled the latent space of a StyleGAN generator and used an attribute classifier to cluster the input space of StyleGAN. This is done based on the probabilities of the classifier. Finally, using the clustered vectors (prototype vector), authors generate images with the desired attributes.

In this work, by employing an autoencoder with a contrastive loss applied to its bottleneck-layer, we were able to model the complex latent space of any StyleGAN generator with a much simpler modeling technique.

\subsection{3D rendering methods} 
Recent advances in computer graphics caused the raise of realistic rendering methods that we often see in the gaming and movie industries. Unfortunately, most of these technologies,
such as \cite{zivadynamic} and \cite{metahuman_whitepaper},
can not be used because of legal restrictions even for research purposes. However, there are some recent works that generate synthetic datasets using 3D rendering pipelines \cite{bae2023digiface, zhang2023dreamface}, but they are not as realistic as their commercial counterparts. One benefit of 3D rendering methods is the access 
to the exact manifold of the models (faces in our case) thus we could easily generate variations of the same identity. As a disadvantage, it is complex to control demographics (e.g., ethnicity) in such methods.
% Also related to our researchers using these kinds of tools to generate synthetic datasets using 3D rendering methods \cite{bae2023digiface}.
% As suggested in \autoref{sec:intro} one can complement our proposed methods with such a method to further create more variations of identities with desired sensitive demographic groups. 

Related to synthetic face dataset generation, authors in \cite{boutros2022sface} trained an identity-conditioned StyleGAN2~\cite{sg2, sg2ada} to alleviate the privacy concerns of current FR datasets.

\subsection{StyleGAN Inversion}\label{sec:ganinversion} 
% StyleGAN inversion is the problem of finding a latent code of an arbitrary image (usually in-domain of the trained network, i.e. If the StyleGAN network trained on the face images, given a face image with similar settings, finding the latent code to produce the input image when passed through synthesis network). More specifically given an input image $\mathbf{I}$ and StyleGAN-based architecture generator $G$ finding the latent that reconstructs the image as close as possible to the input image $\mathbf{I}$, Mostly inversion methods being categorized using the latent space in which they trying to map the input image (e.g. $\mathcal{W}$, $\mathcal{W}^{+}$, $\mathcal{P}$, $\mathcal{S}$, ...) and the method which used to convert the image to the desired space (e.g. Optimization-based, Encoder-based, ... ), for a detailed survey on GAN Inversion interested reader may refer to \cite{xia_gan_2022}.
StyleGAN inversion is the problem of finding the latent code of an arbitrary image, typically within the domain of the trained network. For example, if the StyleGAN network is trained on face images, the task involves finding the latent code that produces the same image when passed through the synthesis network with similar settings. More specifically, given an input image $i$ and a StyleGAN-based 
 generator $G$, the goal is to find the latent code that can reconstruct the input image as closely as possible. Inversion methods are generally defined by: (i) latent space in which they map the input image, spaces such as $\mathcal{W}$, $\mathcal{W}^{+}$, $\mathcal{P}$, $\mathcal{S}$ and (ii) the method used to convert the image to the desired space, such as optimization-based or encoder-based methods. For a more detailed survey of GAN inversion, interested readers may refer to \cite{xia_gan_2022}.

Here we briefly describe the types of StyleGAN inversion methods proposed in the literature, and we show that not all of these methods are suitable for our application in mind.

\begin{description}

\item[Optimization-based]: Most of the optimization-based inversion methods change the weights of the synthesis network for each image. As one of the popular methods \cite{roich_pivotal_2021} optimizes the weights of the generators for each image to steer it to a more editable part of the latent space. In this case, we can not reliably model the latent space since the synthesis network would be different for each image.

\item[HyperNetwork-based]: The benefits of inversion methods such as those found in \cite{wang_high-fidelity_2022} and \cite{dinh_hyperinverter_2022} largely stem from weight correction to the generator using an auxiliary network called hypernetwork. This correction is performed based on a per-image-basis, meaning that the original image or its weight correction is required at sampling time. This prevents the synthesis of images solely from the latent space of the generator.

\item[Encoder-based]: This type of StyleGAN inversion involves using an auxiliary mapping network to convert the input image to the desired latent space (e.g., $\mathcal{W}$, $\mathcal{S}$, $\mathcal{W}^{+}$, $\mathcal{P}$ spaces). This includes various techniques depending on the architecture and final latent space of auxiliary networks. The two most renowned methods in this category are \cite{psp} and \cite{e4e}.

\end{description}

Our key assumption is that the demographics of an image will remain unchanged after inverting it into the desired latent space and reconstructing it using StyleGAN's generator. To verify this assumption, we conducted a qualitative comparison of reconstructed images obtained from the inversion process in the \autoref{sec:experiments}. 
We conclude that the encoder-based inversion method described in this section is the optimal method for our application. In particular, we used the pixel2style2pixel (pSp) \cite{psp} encoder-based inversion method, due to its superior quality compared to \cite{e4e}.

% There is another aspect of different stylegan architecture which that requires our attention, especially with our main purpose of this research in mind, i.e. complementing existing dataset to get the balanced and fair version of them, this requires that the distribution of the generated images be similar to the original datasets.
% Following studies on the domain-gap
% \cite{sg3}
% \cite{frank_leveraging_2020}
% \cite{dong_think_2022}
% especially in the frequency content of generated images from various StyleGAN architectures 
% \cite{sg3}
% \cite{sg2} 
% we decided to employ the StyleGANv3 \cite{sg3} generation methodology for our synthesis process.

% In light of our main research goal, which is to complement existing datasets to create a balanced and fair version of them, we must also consider an important aspect of the different StyleGAN architectures. Specifically, the distribution of the generated images must be similar to that of the original datasets. Several studies, such as \cite{sg3}, \cite{frank_leveraging_2020}, and \cite{dong_think_2022}, have investigated domain-gap issues, particularly in the frequency content of generated images from various StyleGAN architectures, thus, we chose to utilize the StyleGANv3 \cite{sg3} generation methodology for our synthesis process.
As mentioned previously, our primary research objective is to supplement current datasets with a balanced and fair version. To accomplish this goal, we must also take into account an essential aspect of the various StyleGAN architectures: the distribution of the generated images closely resemble that of the original datasets. Several studies, \cite{sg3, frank_leveraging_2020, dong_think_2022}, have investigated the domain-gap issues that arise in the frequency content of generated images produced by different StyleGAN architectures. As it is shown, the StyleGANv3~\cite{sg3} generation method is less prone to this problem. Thus we conclude employing this method for our synthesis process.
%Consequently, we have decided to employ the StyleGANv3 \cite{sg3} generation methodology for our synthesis process.

%%%%%%%% Methodologyw
\section{Proposed Method} \label{sec:proposed_method}

\subsection{Problem Setup}
\vspace{-5pt}
Assume that we have an image dataset $\mathcal{D}$ with domain $d$ (e.g. human face images or animal images) depicted by set $\{\mathcal{D}, d\}$ with demographic groups set $\mathcal{A}$. $\mathcal{A}$ can be defined as $\{\mathcal{A}_{gender}, \mathcal{A}_{race}, \mathcal{A}_{age-group}, ... \}$ in which each of them will take some discrete values (e.g. for $\mathcal{A}_{gender}$ this could be $male$ and $female$ and for $\mathcal{A}_{age-group}$ could be children between age $9$ to $14$ or young adults between age $18$ to $30$ ). Given a StyleGAN generative model, $\mathcal{G}$, trained on the same domain $d$ as in $\mathcal{D}$, our goal here is to model the arbitrary sampling spaces of trained StyleGAN model for being able to generate any combination of demographic groups that were presented in $\mathcal{D}$. As an example, for our FR dataset, we want to generate as many synthetic images of  \emph{hispanic} \emph{male} in his \emph{youth (18-30)} as we want. As mentioned previously, by doing so, our goal would be to alleviate the bias introduced due to the disparity of demographics in current face recognition datasets. Here we limit our experiments to the human faces, the same approach also can be used for any $\{\mathcal{D},d\}$ and StyleGAN generator $\mathcal{G}$ trained on domain $d$.
\autoref{figure:overall_pipeline} illustrates the complete architecture of our generation pipeline for training and inference. Starting from a dataset with demographic, labels such as MORPH \cite{ricanek_morph_2006}, UTKFace \cite{utk20k} or FairFace \cite{karkkainen_fairface_2021} with the images of human faces, we first invert the images using StyleGAN inversion that was trained for $\mathcal{G}$ (described in \autoref{sec:ganinversion}). More specifically given images of $\mathcal{D}$ as $\mathbf{i}$ and inversion network $\mathcal{I}_{\mathcal{G}}$ we compute the inverted latent code, $\mathbf{w}_j$, as: 

\begin{equation}
    \forall j \in \{ 1 , ...  , | \mathcal{D} | \} ;    \mathbf{w}_j = \mathcal{I}_{\mathcal{G}}( \mathbf{i}_j )
    \label{eq:inversion_images}
\end{equation}

Directly modeling the latent space of StyleGANs (e.g., $\mathcal{W}^{+}$) is impossible because it forms an entangled representation (i.e., latent dimensions do not control a single demographic). Therefore, we form an auxiliary space to 
disentangle the representation and hence allow for modeling of this new latent space.
We build this auxiliary space using the bottleneck layer of an autoencoder. 
Finally, by sampling the models according to a specific demographic group (e.g., \emph{white}-\emph{female} or \emph{hispanic}-\emph{man} in his \emph{30s}) and passing the sampled latent space to our networks, we were able to generate synthetic datasets with any specific attribute.
%to complement and reduce the demographic biases that exist in current FR datasets.

%%%%%% Latent Space Modeling
\subsection{Latent Modeling}\label{sec:latent_modeling}
\begin{figure}
    \includegraphics[width=8cm]{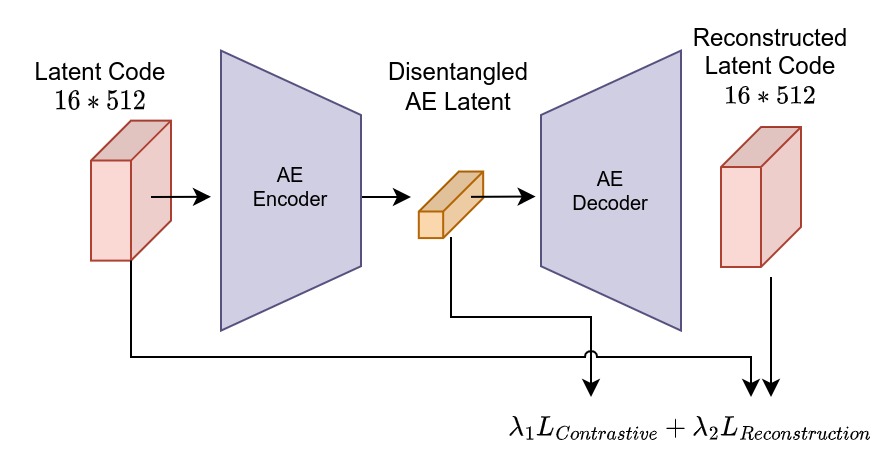}
    \caption{Disentangling latent space using autoencoder with Contrastive Loss}
    \label{fig:autoencoder_disentg} 
    \vspace{-15pt}
\end{figure}

We first explored the possibility of modeling the $\mathcal{W}^{+}$ space (i.e. the output of StyleGAN's inversion~\cite{psp}).
However, as reported in \cite{shukor_semantic_2022} and confirmed by our findings in \autoref{sec:experiments}, this latent space
%aligned with our findings and literature \cite{shukor_semantic_2022}
is too complex for being able to model it directly using either bijective transforms ( i.e. normalizing flows ) or other statistical modeling schemes like GMMs. To alleviate this complexity, we employ an autoencoder network. We denote it in \autoref{fig:autoencoder_disentg}. More specifically: 

\begin{equation}
    \begin{aligned}
    & \mathbf{b} = E (\mathbf{w}), \\
    & \mathbf{w}^{*} = D ( \mathbf{b} ) 
    & \text{where:} \quad \mathbf{w}^{*} \simeq \mathbf{w}. \\ 
    \end{aligned}
\end{equation}
%Where, $ \mathbf{w}^{*} \simeq \mathbf{w} $  

Here $E$ and $D$ are the encoder and decoder parts of the autoencoder respectively, $\mathbf{b}$ is the bottleneck output of the autoencoder (i.e. output of $E$). To ensure the $\mathbf{w}^{*} \simeq \mathbf{w}$ we employ an Euclidean loss between the output of $D$ and input of $E$ ($\mathcal{L}_{Reconstruction}$). 
To enforce the disentanglement of the sensitive demographic groups we employed a contrastive loss applied to the bottleneck layer of autoencoder.
For the contrastive loss, we used the \textbf{LiftedStructured} loss proposed in 
\cite{liftedstrcutred16} 
defined as follows: 

\vspace{-5pt}
\begin{equation}
    \mathcal{L}_{Contrastive} = \frac{1}{2|\mathcal{P}|} \sum_{(i,j) \in \mathcal{P}} \max (0, \mathcal{L}_{i,j}) ^ 2 
    \label{eg:lift_definition}
\vspace{-5pt}
\end{equation}
where, $\mathcal{P}$ is the set of positive samples in the mini-batch and the $\mathcal{L}_{i,j}$ is defined as follows: 
\vspace{-5pt}
\begin{equation}
    \mathcal{L}_{i,j} = \log ( \sum_{ (i,k) \in \mathcal{N} } \exp( \alpha - l_{i,k} )+  \sum_{ (j,l) \in \mathcal{N} } \exp( \alpha - l_{j,l} ) ) + l_{i,j}
\vspace{-5pt}
\end{equation}
Here, the function $l_{m,n}$ is a distance function between $m$-th and $n$-th samples. We set it as Euclidean distance. $\mathcal{N}$ is the set of negative samples in our mini-batch, and $\alpha$ is the negative margin.
By applying contrastive loss on different demographic groups separately, our overall contrastive loss will be the combination of each loss for each demographic group as follows :  
\vspace{-5pt}
\begin{equation}
    \mathcal{L}_{Contrastive}^{Total} = \sum_{g \in \mathcal{A}} c_{g} \mathcal{L}_{Contrastive}^{g} 
    \label{eq:total_contrastive}
    \vspace{-5pt}
\end{equation}
Here, $g \in \mathcal{A}$ means that the contrastive loss is applied to either of $\{ \mathcal{A}_{gender}, \mathcal{A}_{race}, \mathcal{A}_{age-group} \}$, separately. In \autoref{eq:total_contrastive}, $c_{g}$ can be used to control the importance of demographic factors (i.e., $\mathcal{L}_{Constrastive}^{g}$).
As mentioned before, for training our autoencoder we also included an Euclidean distance as our reconstruction loss between the $\mathbf{w}$ and $\mathbf{w^{*}}$, so the total loss will be the weighted sum of reconstruction and contrastive loss as follows: 
\begin{equation}
    \mathcal{L}_{Total} = \lambda_{1}\mathcal{L}_{Contrastive}^{Total} + \lambda_{2}\mathcal{L}_{Reconstruction}
    \label{eq:total_loss_autoencoder}
    \vspace{-5pt}
\end{equation}

In \autoref{eq:total_loss_autoencoder}, $\lambda_{1}$ and $\lambda_{2}$ are to control the contribution of contrastive and reconstruction loss respectively. 
\begin{figure*}
    \includegraphics[width=\textwidth]{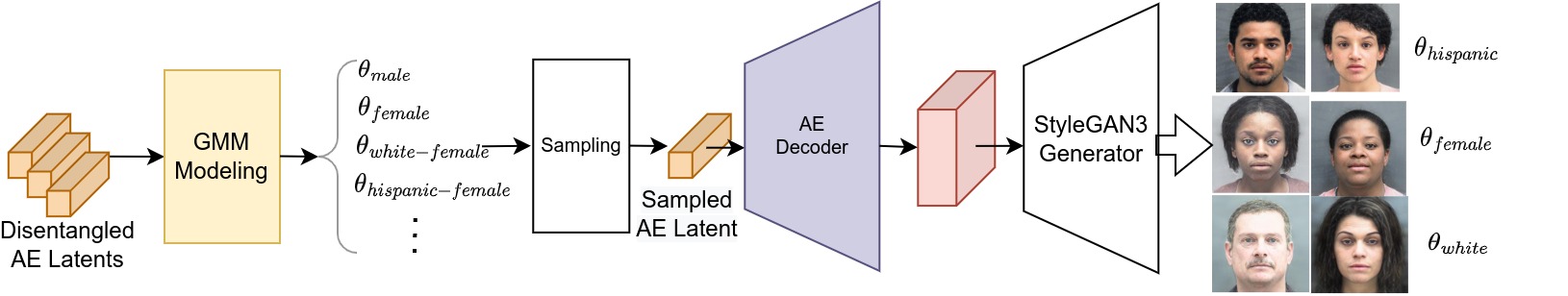}
    \caption{AE-Latent Disentangled Modeling and Sampling}
    \label{fig:modeling_sampling}
    \vspace{-8pt}
\end{figure*}
\subsection{Gaussian Mixture Modeling}
Assuming a disentangled space (i.e. $\mathbf{b}$), we can employ traditional techniques such as Gaussian Mixture Models (GMM)~\cite{gmm2009} which is defined as follows: 
 \vspace{-5pt}
\begin{equation} 
    \begin{aligned}
    \mathcal{M} ( \mathbf{b} ; \theta_{g}) & =  \sum_{m=1}^{M} w_m \mathcal{N}(\mathbf{b} | \mathbf{\mu}_{m}, \mathbf{\Sigma}_{m}) \\
    %\mathcal{N}(\mathbf{b} | \mathbf{\mu}_{i}, \mathbf{\Sigma}_{i}) & = \frac{1}{(2\pi)^{D/2}|\Sigma_i|^{1/2}} e^{-\frac{1}{2} ({\mathbf{b} - \mathbf{\mu}_i})^{T} {\mathbf{\Sigma}_i}^{-1}  ({\mathbf{b} - \mathbf{\mu}_i}) }
    \end{aligned}
    \vspace{-5pt}
\end{equation}

Here, $\mathcal{N}(\mathbf{b}|  \mathbf{\mu}_{m}, \mathbf{\Sigma}_{m})$ is the multivariate Gaussian distribution, $M$ is the number of mixture components, $\mathbf{\mu}_m$, $\mathbf{\Sigma_m}$ and $w_m$ are mean, covariance matrix and weight of mixture component number $m$ respectively. The weights must satisfy $\sum_{m=1}^{M} w_m = 1$. $\theta_g$ is a set of all the mentioned parameters. 
% \autoref{fig:modeling_sampling} illustrates this process.
To model the space for a given demographic group, $g$, we use the Expectation-maximization~\cite{moon1996expectation} algorithm on the samples in $g$ demographic group to solve for parameters $\theta_g$.
 As an example, we fit a GMM to the $male$ group and another one for a $hispanic-female$ demographic, respectively denoted by $ \mathcal{M}(\mathbf{b}; \theta_{male}$) and  $ \mathcal{M}(\mathbf{b}; \theta_{hispanic-female})$.  Here we can compute the likelihood of a sample being drawn from $g$ as $P(\mathbf{b} | \theta_g)$, like-wise, the log-likelihood (LL) can be formulated as $LL = log ( P(\mathbf{b} | \theta_g) )$.

\subsection{Generating Images with the proposed approach} 
% As depicted in \autoref{fig:modeling_sampling}, by sampling the desired demographic groups according to its GMM parameters (i.e. $\theta_{i}$ s) given the bottleneck of our autoencoder, later on bypassing the vectors to the decoder part of our autoencoder we get tensors supposedly presents desired demographic groups in the $\mathcal{W}^{+}$ space of the StyleGAN3's latent space, later on by passing these tensor to generator part of the StyleGAN3 we get the face images according to a desired sampled demographic group of GMMs, this process also can be viewed in \autoref{alg:gen_demo}.
As shown in \autoref{fig:modeling_sampling}, we first sample $\mathbf{b}$ according to desired demographic groups by using their corresponding GMM parameters (i.e., $\theta_{g}$). Then, we use  decoder part of our autoencoder ($D(\mathbf{b})$) to obtain latent code that represents the desired demographic groups in the latent space of interest (e.g. $\mathcal{W}^{+}$ latent space of StyleGANv3). Finally, we pass these latent codes to the StyleGAN's generator to obtain face images that correspond to the desired demographic group sampled from the GMMs. This process is illustrated in Algorithm \autoref{alg:gen_demo}.

\begin{algorithm}
\caption{Generating images of desired demographic}\label{alg:gen_demo}

\hspace*{\algorithmicindent} \textbf{Input}: $\mathcal{G}$, $D$, $\theta_{g}$ \\
\hspace*{\algorithmicindent} \textbf{Output}: $\mathbf{i}_{g}$
\begin{algorithmic}
    \State $ \mathbf{b} \sim \mathcal{M}(\mathbf{b}; \theta_{g})  $: Calculating latent according to desired demographic  
    \State $ \mathbf{i}_{g} \gets \mathcal{G} ( D ( \mathbf{b} ) )  $: Generating image from the latent
\end{algorithmic}
\end{algorithm}

%%%%%% Experiments 
 \section{Experiments} \label{sec:experiments}
In this section, we describe our setup, implementation details, and various experiments that we employ to validate our results. 

\subsection{Validation of Synthesis}\label{sec:valid_class_synthe}
To determine if the generated images are following the desired demographic (i.e. $g$ in $\mathbf{i}_{g}$ in algorithm \ref{alg:gen_demo}), we employed an image classification task. We used the fair classifier model provided by the \cite{karkkainen_fairface_2021}. We used the MORPH dataset for training our autoencoders. Thus the $\mathcal{G}$ which was trained on FFHQ \cite{sg1} and our autoencoder that trained on MORPH did not have any prior exposure to the images used to train the FairFace classifier.
%The \cite{genprompt22, shukor_semantic_2022} have similar application as our technique. 
%In case of \cite{shukor_semantic_2022} they did not release their code.  We also could not compare with \cite{genprompt22}, as their provided code uses the FairFace classifier in their generation stage (as supervision), which is the classifier that we used for verification of our results.

\begin{figure}[!htb]
   % \begin{minipage}{0.5\textwidth}
   %   \centering
   %   \includegraphics[width=\linewidth]{imgs/GenOurs.png}
   %   \caption{Interpolation for Data 1}\label{fig:ours_race_classification_fairface}
   % \end{minipage}
   % \begin{minipage}{0.5\textwidth}
   %   \centering
   %   \includegraphics[width=\linewidth]{imgs/GenGenerativePrompt.png}
   %   \caption{Interpolation for Data 2}\label{fig:genprompt_race_classification_fairface}
   % \end{minipage}

   \centering
   \begin{subfigure}[b]{0.55\linewidth}
       \centering
       \includegraphics[width=\linewidth]{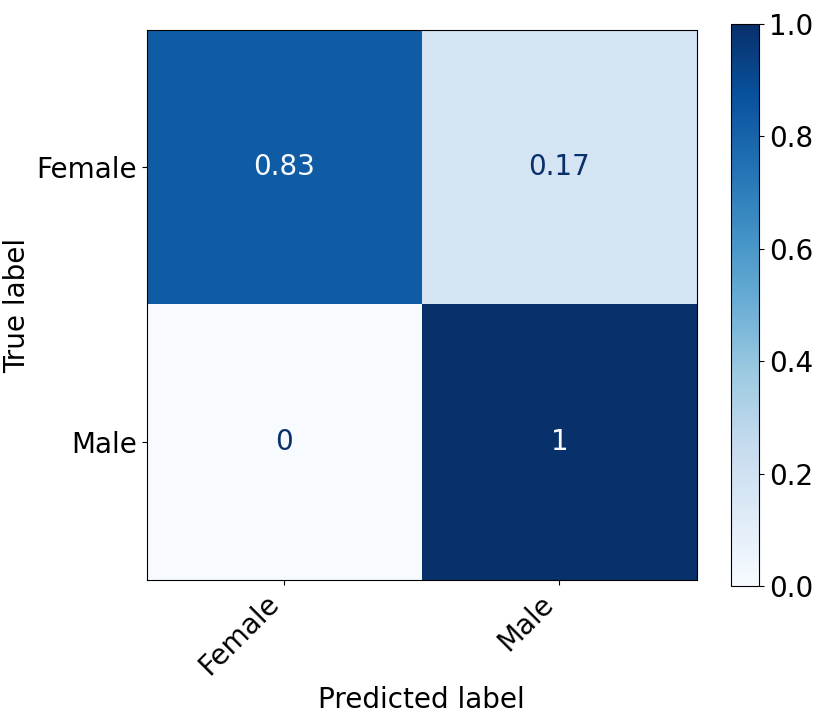}
   \end{subfigure}%
   % \begin{subfigure}[b]{0.5\linewidth}
   %     \centering
   %     \includegraphics[width=\linewidth]{imgs/GenGenerativePrompt.png}
   % \end{subfigure}
   \caption{Confusion matrix of the gender classification task for generated images using fair classifier model }
    \label{fig:gender_classificatoin_fairface_confusion_matrix}
    \vspace{-10pt}
\end{figure}

\begin{figure}[!htb]
    \centering
    \begin{subfigure}[b]{0.85\linewidth}
        \centering
        \includegraphics[width=\linewidth]{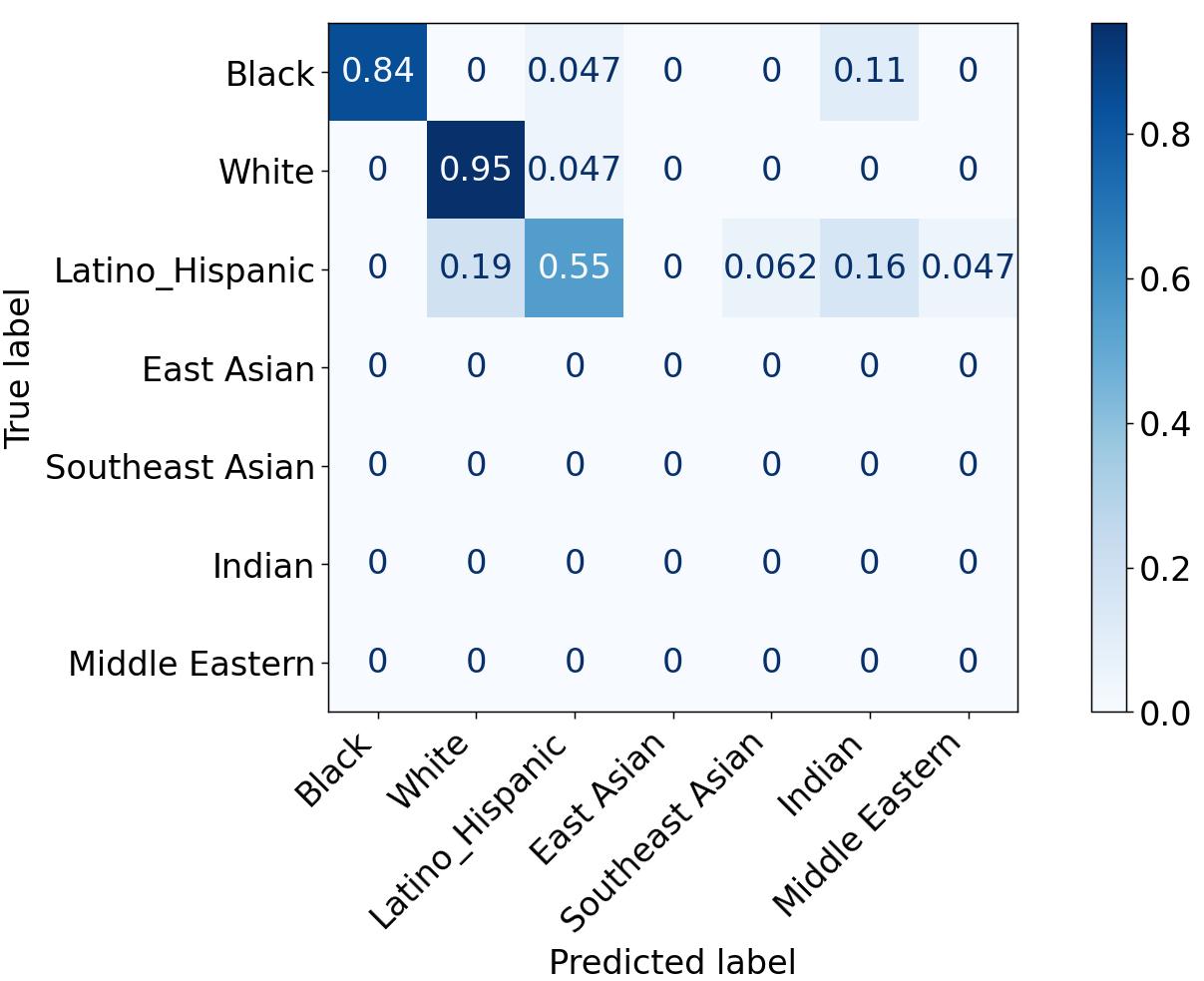}
    \end{subfigure}%
    % \begin{subfigure}[b]{0.5\linewidth}
    %     \centering
    %     \includegraphics[width=\linewidth]{imgs/RaceGenPrompt.png}
    % \end{subfigure}
   %\subfigure[]{\includegraphics[width=1\linewidth]{}}
   %\subfigure[]{\includegraphics[width=1\linewidth]%{imgs/RaceGenPrompt.png}}
   \caption{Confusion matrix of the race classification task for generated images using fair classifier model}
    \label{fig:race_classificatoin_fairface_confusion_matrix}
    \vspace{-10pt}
\end{figure}

\autoref{fig:gender_classificatoin_fairface_confusion_matrix} and \autoref{fig:race_classificatoin_fairface_confusion_matrix} shows the confusion matrix for gender and race classification respectively. Using our method we generate \emph{1000} image for each \emph{male} and \emph{female} and perform the gender classification. 
For race classification, we did the same with \emph{White}, \emph{Black} and \emph{Latino-Hispanic}. We did not include \emph{Asian} race demographic in this experiment as the number of samples of the MORPH dataset which we trained our autoencoder on them was to small. Also, note that the MORPH dataset only has $5$ demographics for race, $\{Black, White, LatinoHispanic, Asian, Unkown\}$. The $7$ classes in \autoref{fig:race_classificatoin_fairface_confusion_matrix} are shown as we used the FairFace classifier. From the figures we can observe that the synthesized face images are following the group that they are sampled from.

%%%%%%%%%%%%%%%%%%%%%%%% CHRISTOPHES EXPERIMENTS %%%%%%%%%%%%%%%%%%%%%%%%%%%
 %% Christophes experiments 

\subsection{Face Recognition Experiments} \label{sec:face_experiments}

Sampling demographic-specific latent $\mathbf{b}$ from a given model $\mathcal{M}(\mathbf{b} ; \theta_{g})$ does not necessarily guarantee that different identities will be generated. To this end, we perform FR experiment on synthesized images to verify two hypotheses: (i) the images created are part of the original data distribution of the MORPH dataset, and (ii) different identities are produced by the proposed method.
The face representation is extracted using a ResNet50 network \cite{HeDeepresiduallearning2016} trained on the WebFace4M dataset \cite{zhu2021webface260m} using the ArcFace loss function \cite{DengArcFaceAdditiveAngular2019a}. Each pair of sample is compared using the similarity function $\mathcal{S}\left(u, v\right) = \frac{u \cdot v}{\|u\|_2 \|v\|_2} - 1$, spanning $\left[-2, 0\right]$. 

\begin{figure}[h!]
    \centering
    \begin{subfigure}[b]{\linewidth}
        \centering
        \includegraphics[width=0.9\linewidth]{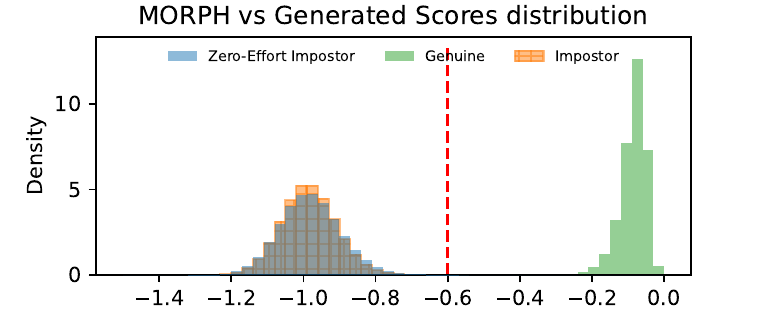}
        \caption{Scores of generated images against the scores of natural images of the original dataset: Genuine pairs (green), Zero-effort Impostors (blue), and Synthetic Impostors (orange).}
        \label{fig:morph_vs_synth_dist}
    \end{subfigure}
    \begin{subfigure}[b]{\linewidth}
        \centering
        \includegraphics[width=0.9\linewidth]{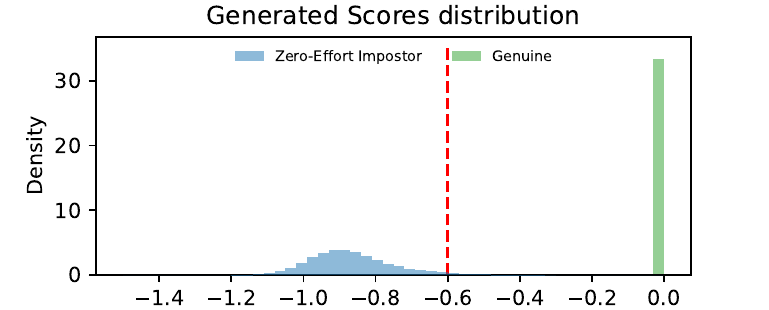}
        \caption{Scores of generated images: Genuine pairs (green), Zero-effort Impostors (blue)}
        \label{fig:morph_synth_dist}
    \end{subfigure}
    \caption{Face recognition scores distributions}
    \label{fig:fr_dist_scores}
    \vspace{-15pt}
\end{figure}

To assess that generated samples are part of the original data distribution, we compare the scores distribution of the natural image of the original MORPH dataset against the synthetically generated samples. \autoref{fig:morph_vs_synth_dist} shows how the synthetic impostors (orange) compare to the real zero-effort impostors scores distribution (blue). The overlap highlights that the sampled images belong to the original data distribution and supports (i). 
With synthetically generated images using the proposed method, it is not possible to compare pairs of images of the same subject, as the sampling scheme does not allow to generate variability (\ie pose, facial expressions, illumination) of a specific face. Therefore we can only compare the synthetic image's zero-effort impostor scores distribution to the original one to assess how different are the generated identities are. 

\autoref{fig:morph_synth_dist} shows how scores change when comparing synthetic images with themselves. The genuine score distribution is represented by a single bin because only a single synthetic image is available per identity. The zero-effort distribution (blue) moves toward the genuine score distribution (green). This shift indicates the identity difference is smaller than in the original dataset. However, the distance between the distributions remains large enough to discriminate between identities.

% - Why we did FR experiments?
%  - ensure we're sampling different identities
% - Process: 
%   - Feature extracted using iresnet50 network trained on WebFace4M dataset
% - Scoring with similarity defined as S(u, v) = u . v / (un * vn) - 1    in [-2, 0]
% 

%%%%%%%%%%%%%%%%%%%%%%%% CHRISTOPHES EXPERIMENTS %%%%%%%%%%%%%%%%%%%%%%%%%%%

\subsection{Demographic Preservation}
\begin{figure}
    \centering
    \includegraphics[width=1\linewidth]{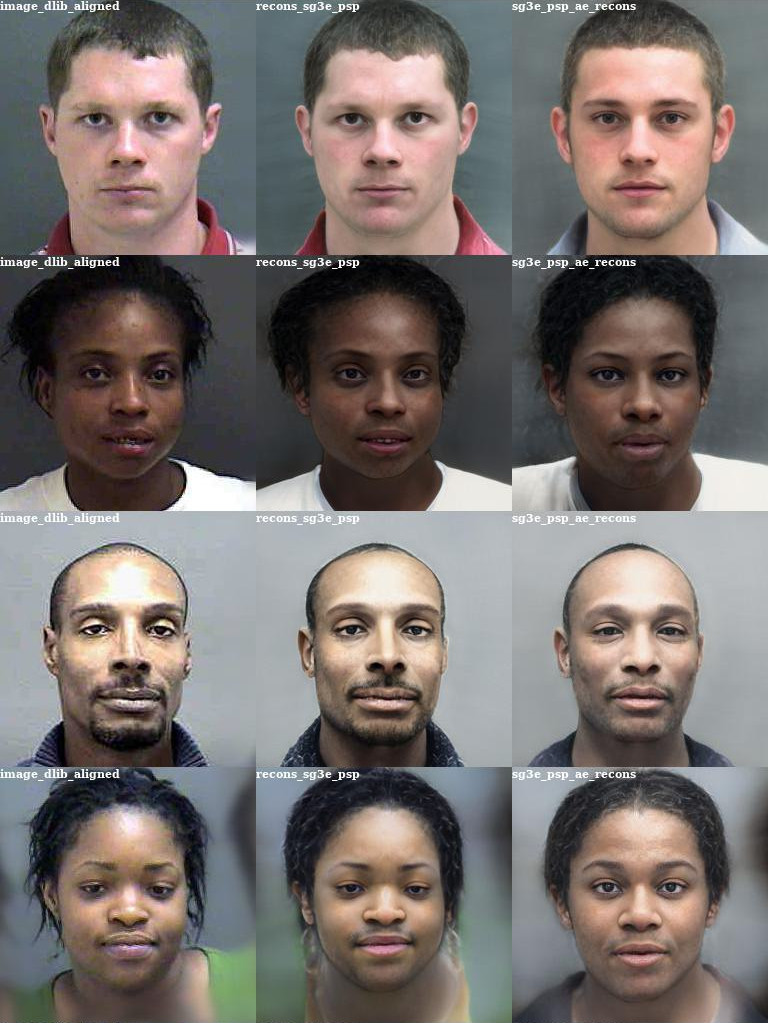}
    \caption{From left to right: the original images of MORPH dataset; reconstruction by the pSp inversion by the StyleGAN3's generator and reconstruction of the pSp inversion when passed through our disentangled autoencoder and later on passed to the StyleGAN3's generator.}
    \label{fig:autoencoder_disentg_recons} 
    \vspace{-10pt}
\end{figure}

\autoref{fig:autoencoder_disentg_recons} shows the reconstruction quality of the result of the pSp and also the reconstruction of the output of our autoencoder, more specifically, second and third columns represent  $\mathcal{G} ( \mathcal{I}_{\mathcal{G}} ( \mathbf{i} ) )$ and $\mathcal{G} ( D ( E ( \mathcal{I}_{\mathcal{G}} ( \mathbf{i} ) ) ) ) $ respectively.  
% \begin{equation}
%     \begin{aligned}
%     \text{second column} &: \mathcal{G} ( \mathcal{I}_{\mathcal{G}} ( i ) ) \\
%     \text{third column} &: \mathcal{G} ( D ( E ( \mathcal{I}_{\mathcal{G}} ( i ) ) ) ) 
%     \end{aligned}
% \end{equation}
Here $\mathbf{i}$ is the original image in the dataset. Qualitatively by comparing columns in \autoref{fig:autoencoder_disentg_recons}, we can observe that although some operations (i.e., contrastive loss) are applied to disentangle the latent space (third column), our demographic groups of interest (e.g., age, gender and race) are preserved. 
\subsection{Latent Space Modeling and Visualization}\label{subsec:observ}
In this section, we show the effectiveness of our disentanglement for modeling the desired latent space. 
\vspace{-5pt}
\subsubsection{t-SNE Visualization}
\begin{figure}
    \begin{subfigure}[b]{0.49\linewidth}
       \centering\captionsetup{width=.8\linewidth}
       \includegraphics[width=\linewidth]{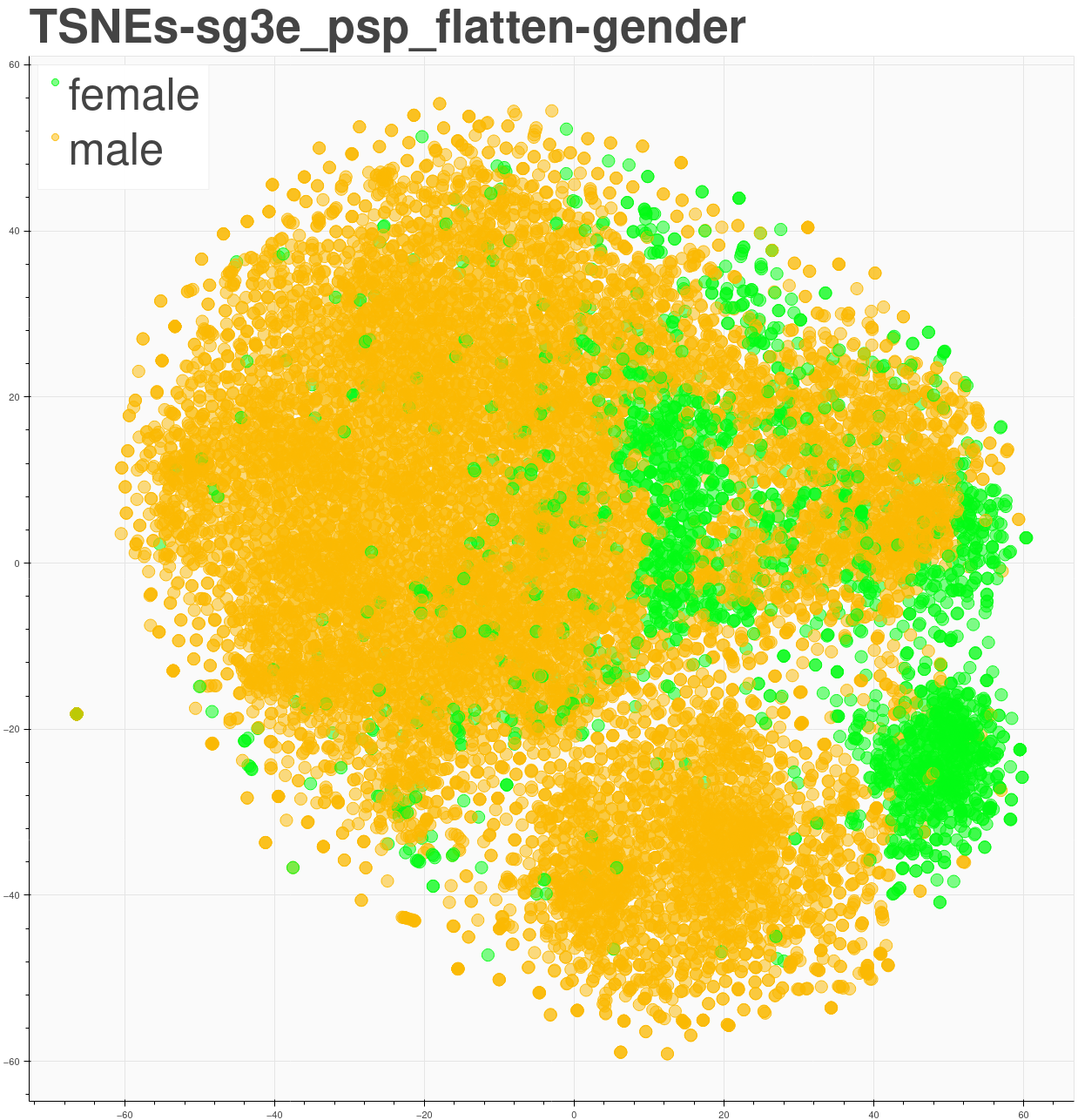}
       \caption{Original $\mathcal{W}^{+}$ space for \emph{gender} demographic.}
    \end{subfigure}
    \begin{subfigure}[b]{0.49\linewidth}
       \centering\captionsetup{width=.8\linewidth}
       \includegraphics[width=\linewidth]{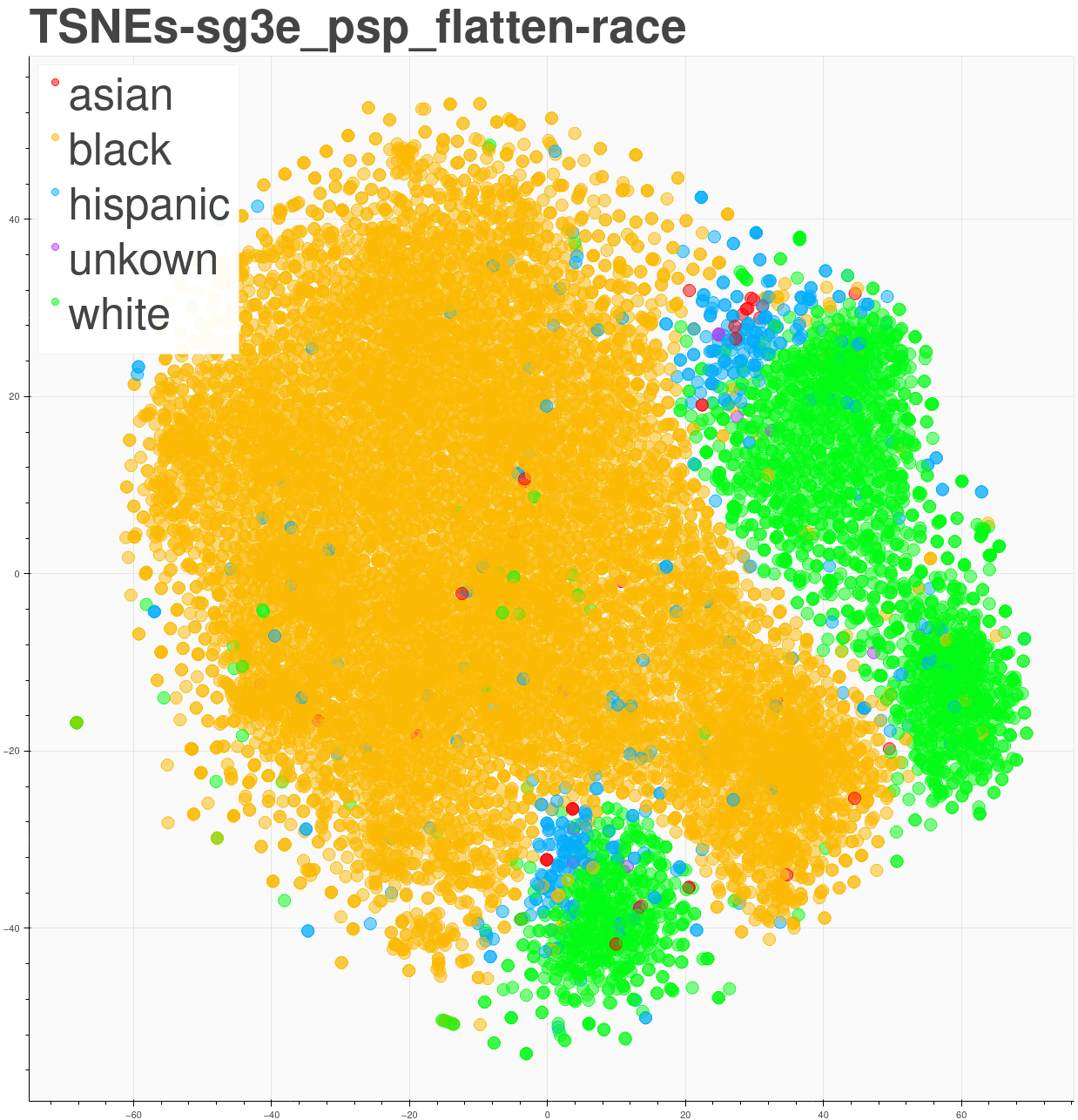}
       \caption{Original $\mathcal{W}^{+}$ space for \emph{race} demographic.}
    \end{subfigure}

    \caption{t-SNE plots for gender and race on the original $\mathcal{W}^{+}$ latent space of the StyleGANv3.}
    \label{fig:tsnes_orig}
    \vspace{-15pt}
\end{figure}

\begin{figure*}
  
   \centering
   %% Using Contrastive 
    \scalebox{0.92}{
   \begin{subfigure}[t]{0.249\linewidth}
       \centering
       \captionsetup{width=.8\linewidth}
       \includegraphics[width=\linewidth]{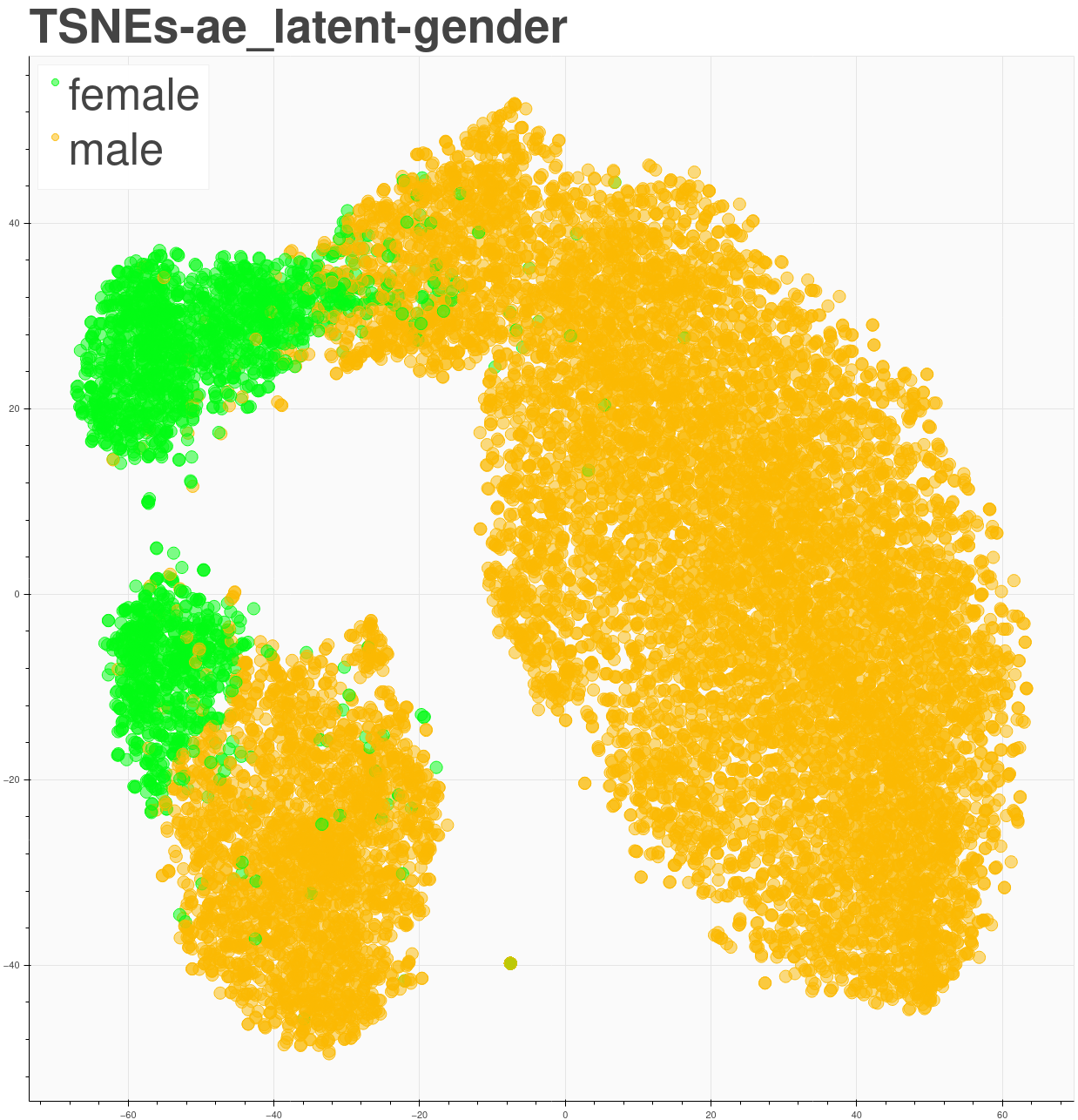}
       \caption{AE's latent space with contrastive loss for \emph{gender}}
   \end{subfigure}%
   \begin{subfigure}[t]{0.249\linewidth}
       \centering
       \captionsetup{width=.8\linewidth}
       \includegraphics[width=\linewidth]{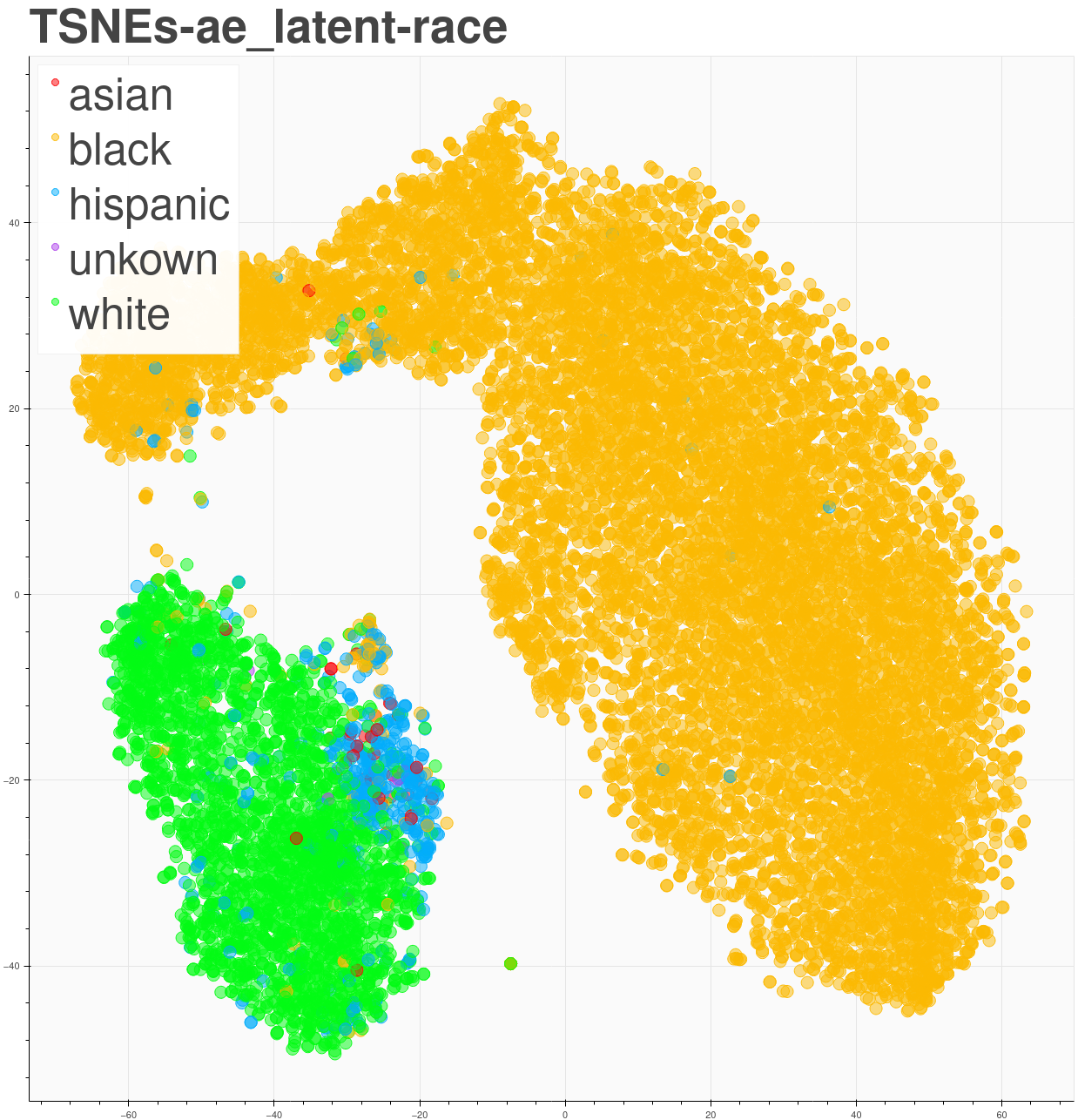}
        \caption{AE's latent space with contrastive loss for \emph{race}}
   \end{subfigure}%
    \begin{subfigure}[t]{0.249\linewidth}
       \centering
       \captionsetup{width=.8\linewidth}
       \includegraphics[width=\linewidth]{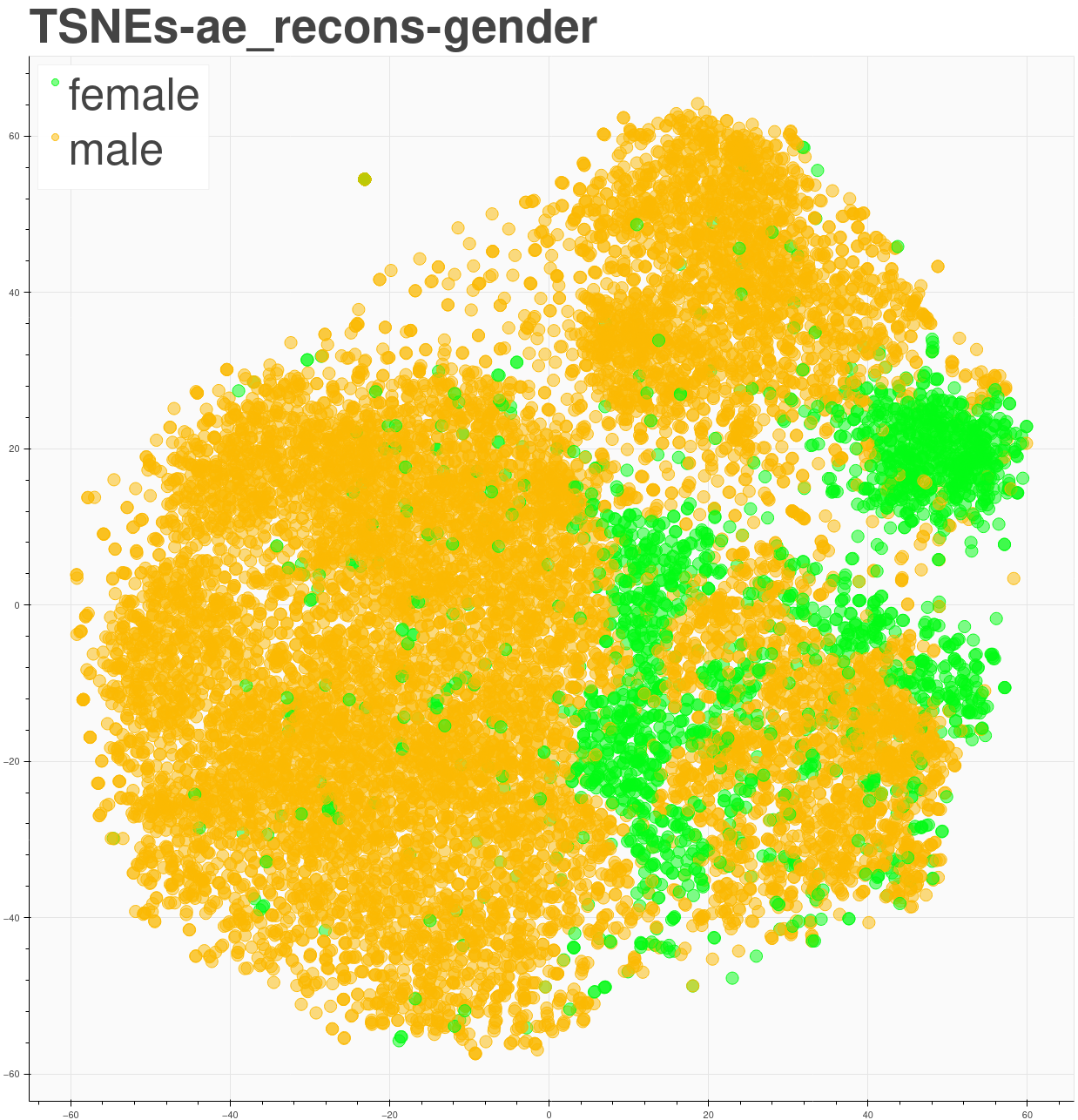}
       \caption{AE's reconstructed space with contrastive loss for \emph{gender}}
   \end{subfigure}%
    \begin{subfigure}[t]{0.249\linewidth}
       \centering
       \captionsetup{width=.8\linewidth}
       \includegraphics[width=\linewidth]{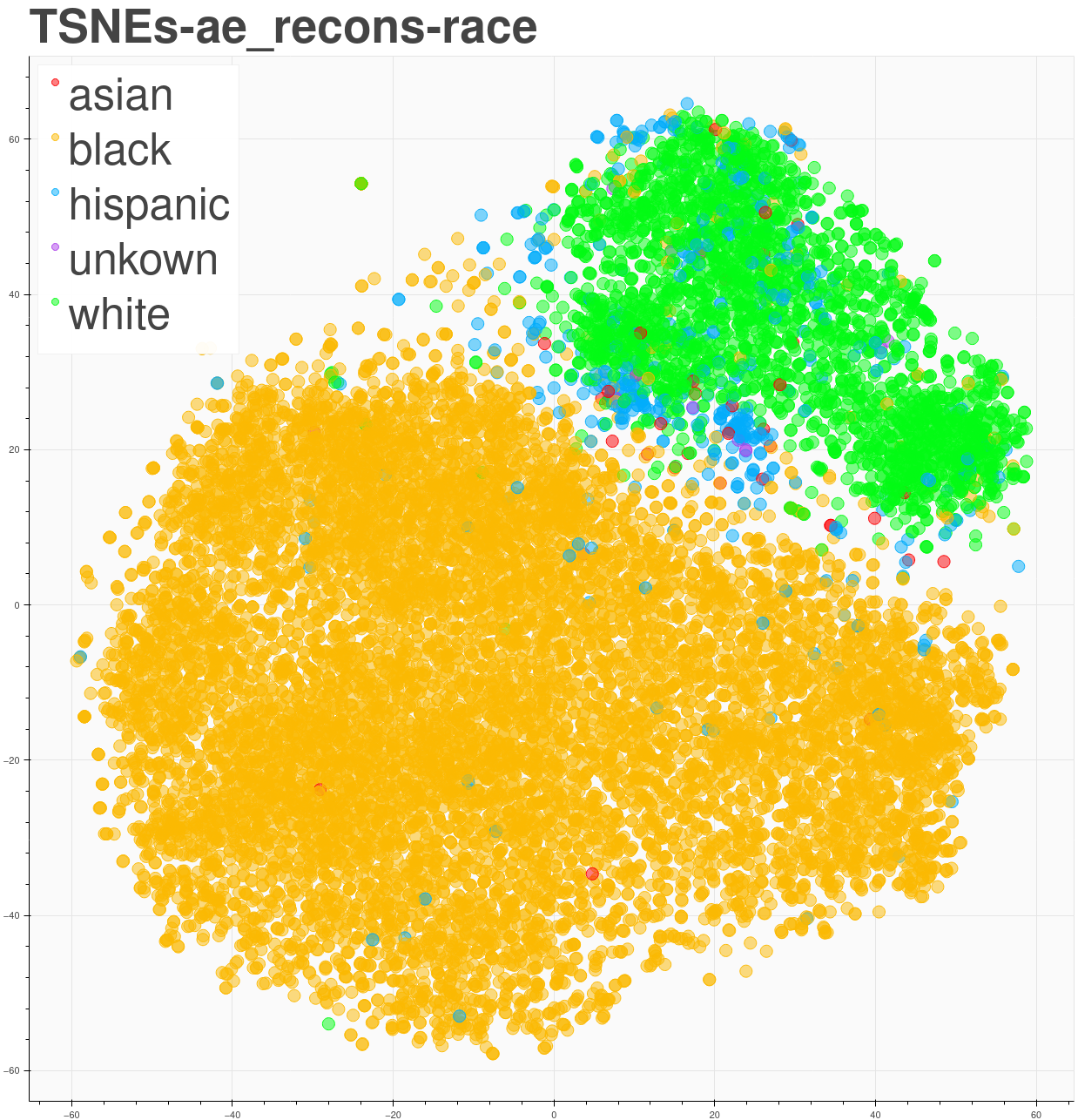}
       \caption{AE's reconstructed space with contrastive loss for \emph{race}}
   \end{subfigure}} \\%

   \scalebox{0.92}{
    \begin{subfigure}[t]{0.249\linewidth}
       \centering
       \captionsetup{width=.8\linewidth}
       \includegraphics[width=\linewidth]{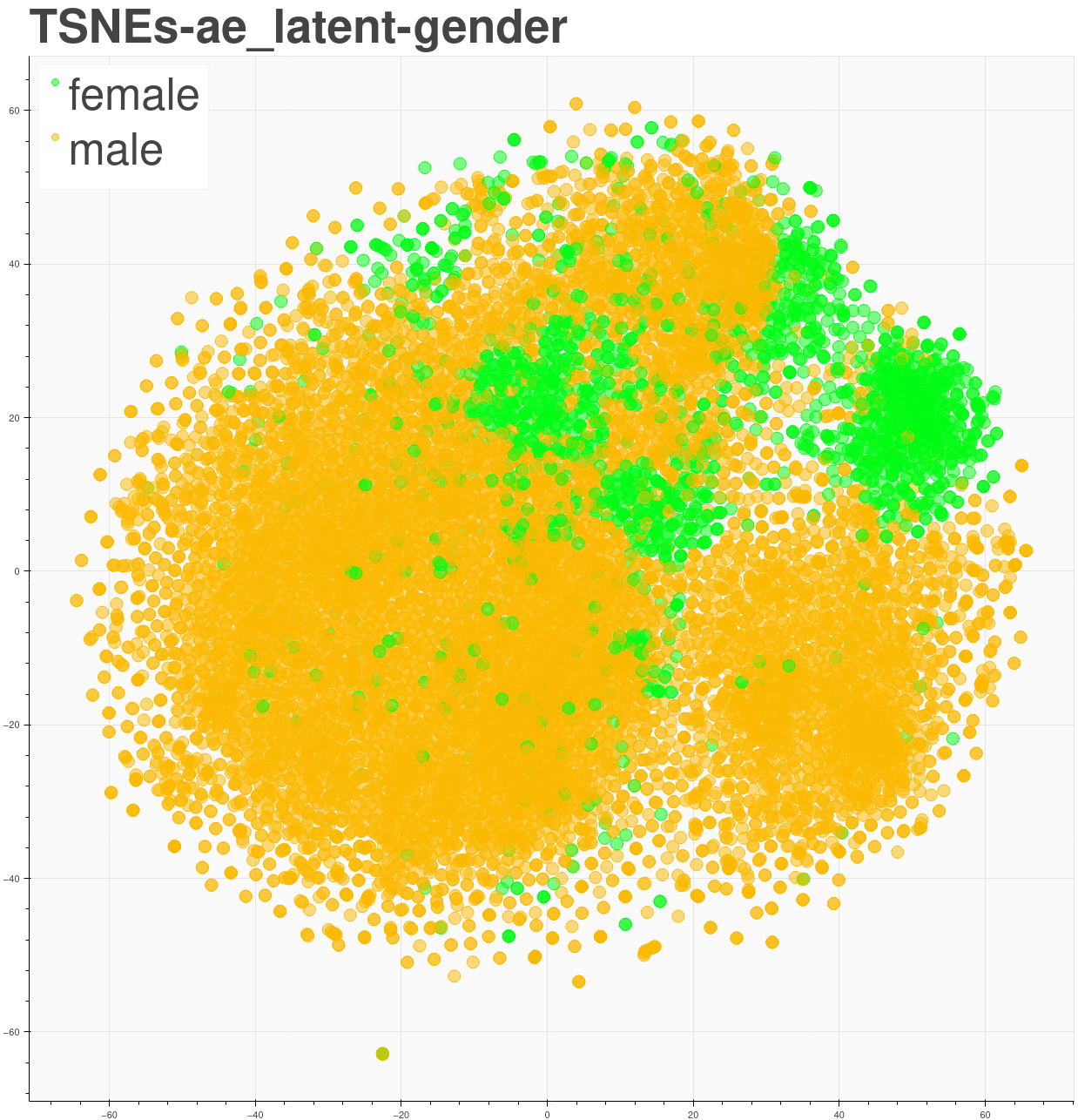}
       \caption{AE's latent space  without contrastive loss for \emph{gender}}
   \end{subfigure}%
   \begin{subfigure}[t]{0.249\linewidth}
       \centering
       \captionsetup{width=.8\linewidth}
       \includegraphics[width=\linewidth]{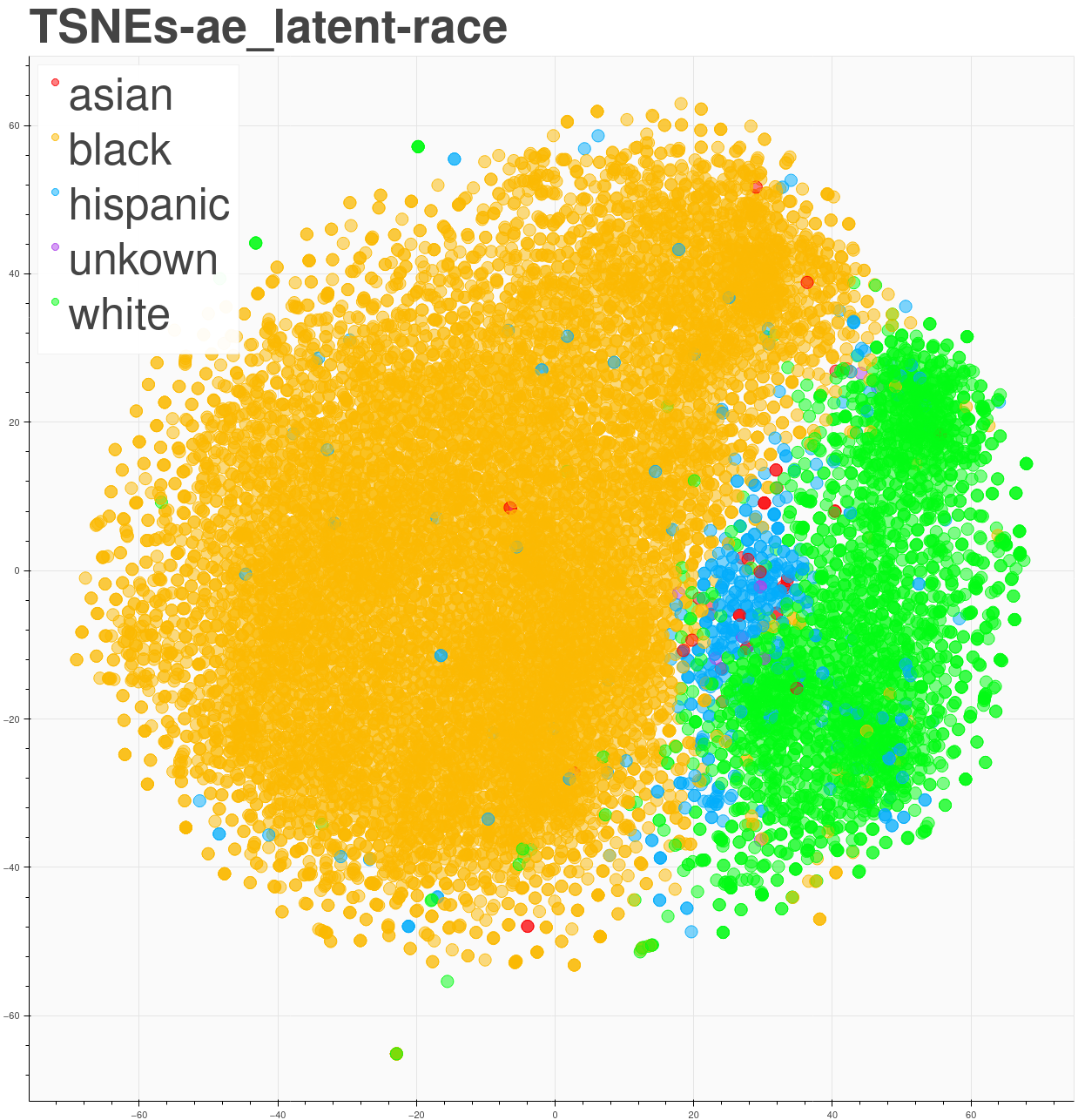}
        \caption{AE's latent space  without contrastive loss for \emph{race}}
   \end{subfigure}%
    \begin{subfigure}[t]{0.249\linewidth}
       \centering
       \captionsetup{width=.8\linewidth}
       \includegraphics[width=\linewidth]{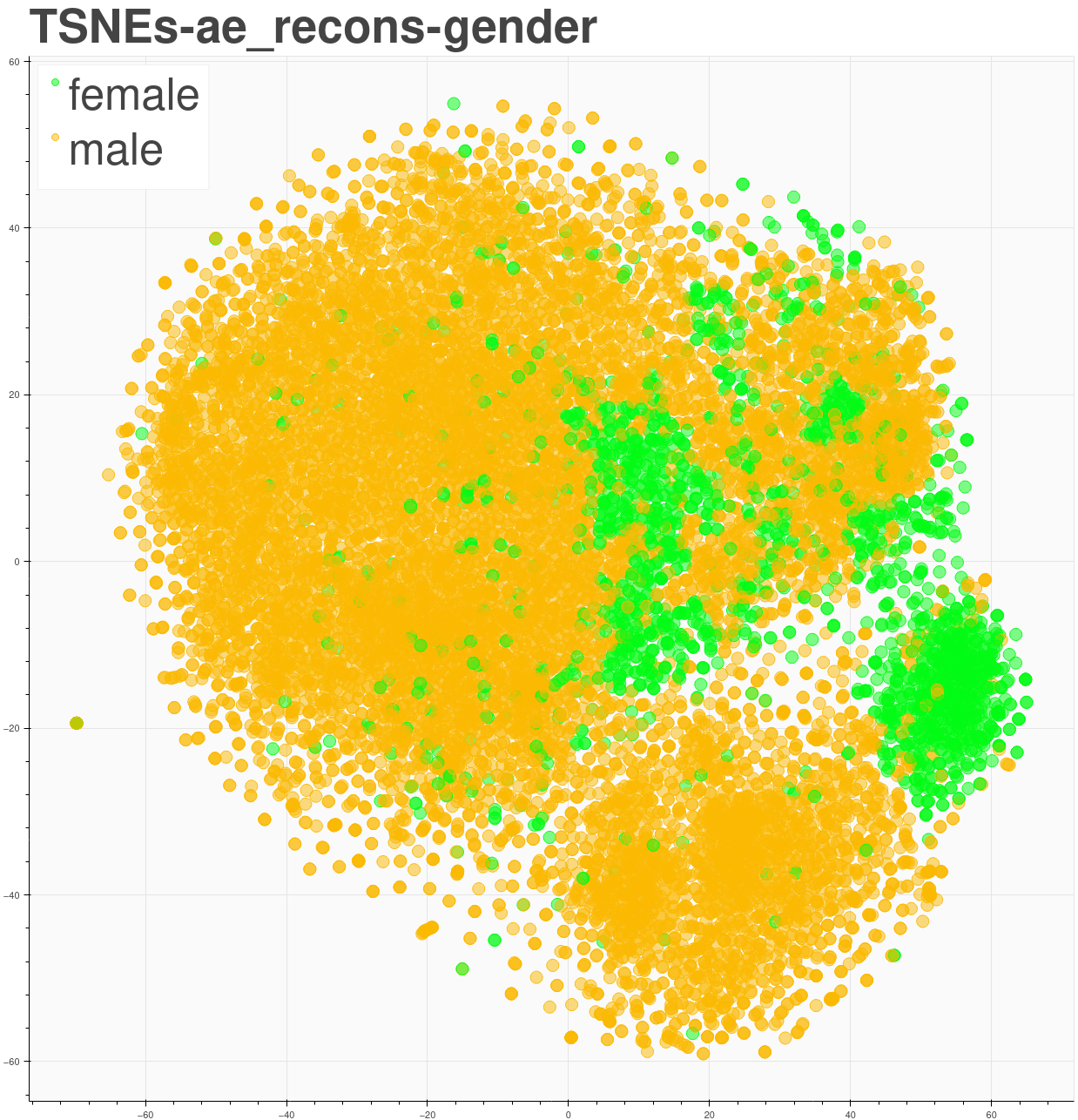}
       \caption{AE's reconstructed space without contrastive loss for \emph{gender}}
    \end{subfigure}%
    \begin{subfigure}[t]{0.249\linewidth}
       \centering
       \captionsetup{width=.8\linewidth}
       \includegraphics[width=\linewidth]{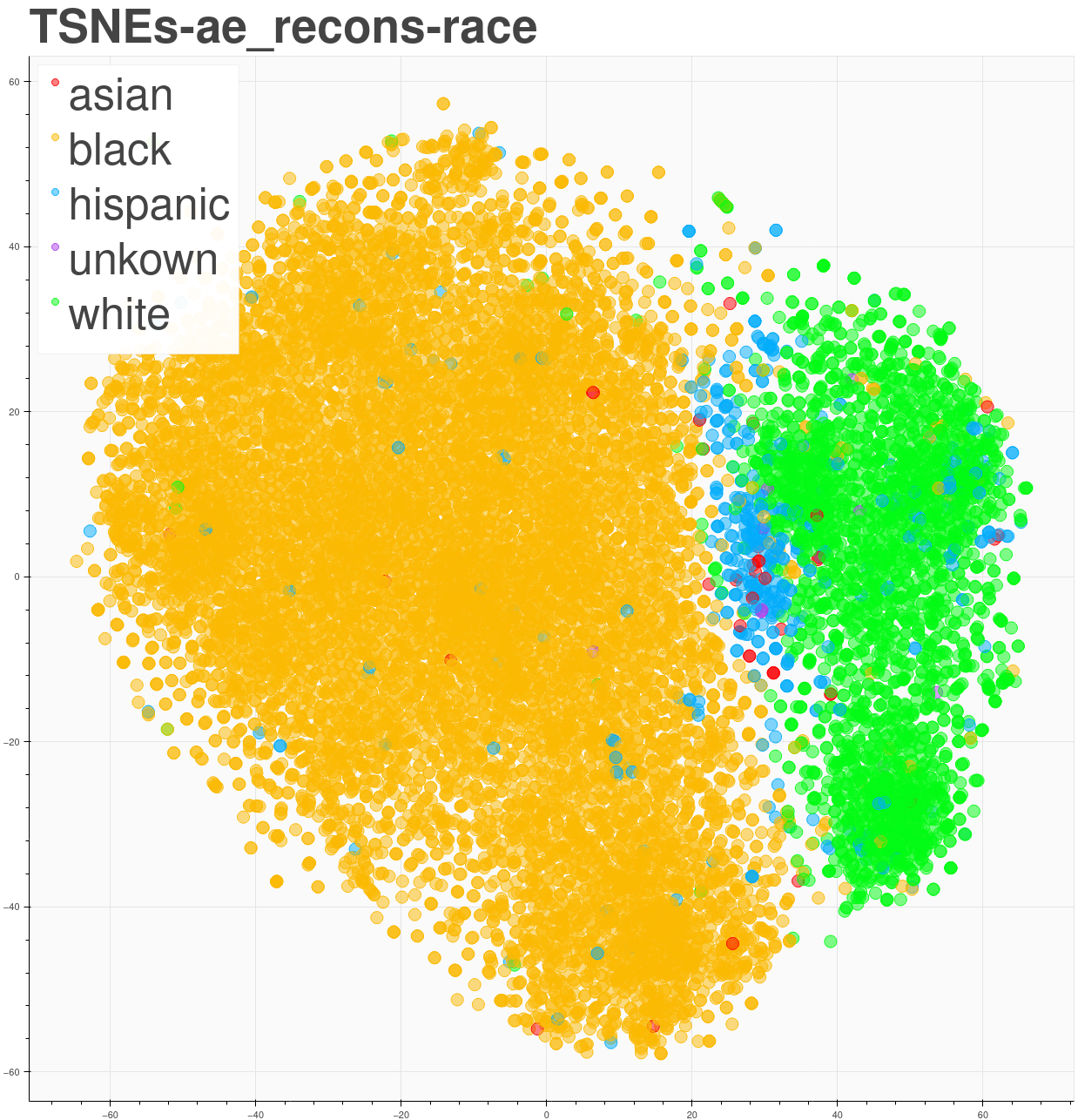}
       \caption{AE's reconstructed space without contrastive loss for \emph{race}}
   \end{subfigure}%
    }
    \caption{t-SNE plots of various latent spaces for test part of the MORPH dataset after learning t-SNE transformation using train split of MORPH. }
    \label{fig:tsnes_ae}
    \vspace{-14pt}
\end{figure*}

\vspace{-5pt}
To visually observe the complex nature for latent space of StyleGAN, we used the t-SNE plots on the test subset of MORPH dataset. In \autoref{fig:tsnes_orig} we visualize the $\mathcal{W}^{+}$ of MORPH according to (a) gender and (b) ethnicity respectively. We can observe that gender and ethnicities according to different values are entangled and complex to model. 
In \autoref{fig:tsnes_ae} we show effectiveness of our disentanglement method on the autoencoder's (AE) bottleneck ( $ \{ E(\mathcal{I}_{\mathcal{G}}(\mathbf{i}_j)) | \mathbf{i}_j \in \mathcal{D}_{test} \}$ ) and reconstruction output $ \{ D(E(\mathcal{I}_{\mathcal{G}}(\mathbf{i}_j))) | \mathbf{i}_j \in \mathcal{D}_{test} \}$ with (a)-(d) and without (e)-(h) our contrastive loss. We can observe that the AE's latent space with the applied contrastive loss is better disentangled according to possible demographics. 


%%% GMM %%%% plots
\begin{figure*}
    \centering
    % \resizebox{.4\totalheight}{!}
    % {

    \scalebox{0.92}{
    \begin{subfigure}[t]{\linewidth}
       \centering
       \captionsetup{width=.8\linewidth}
       \includegraphics[width=\linewidth]{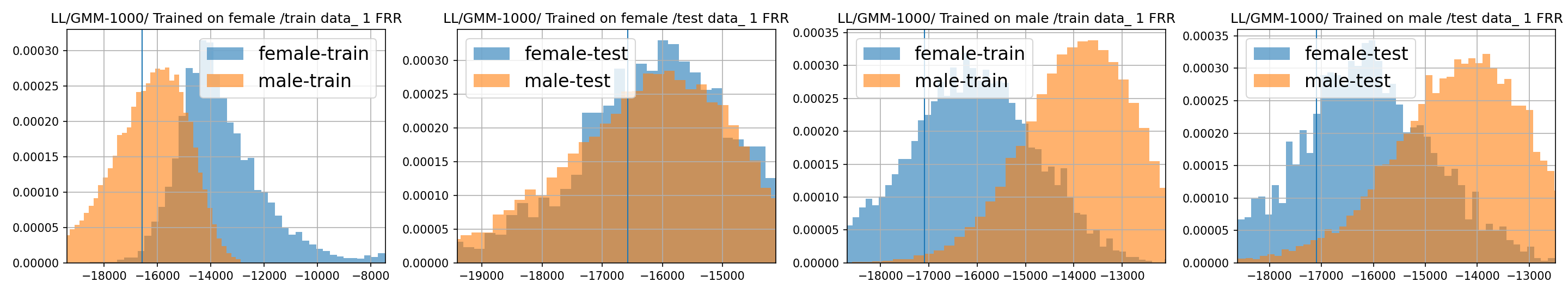}
       \caption{Trained on the StyleGANs original $\mathcal{W}^{+}$ space }
    \end{subfigure}}\\
   %  \scalebox{0.92}{
   %  \begin{subfigure}[t]{\linewidth}
   %     \centering
   %     \captionsetup{width=.8\linewidth}
   %     \includegraphics[width=\linewidth]{imgs/ll_plots/wo_cont_ll_compairs_plot_ae_latent.png}
   %     \caption{GMMs trained on AEs latent space that itself trained without any contrastive loss}
   % \end{subfigure}}\\
    \scalebox{0.92}{
    \begin{subfigure}[t]{\linewidth}
       \centering
       \captionsetup{width=.8\linewidth}
       \includegraphics[width=\linewidth]{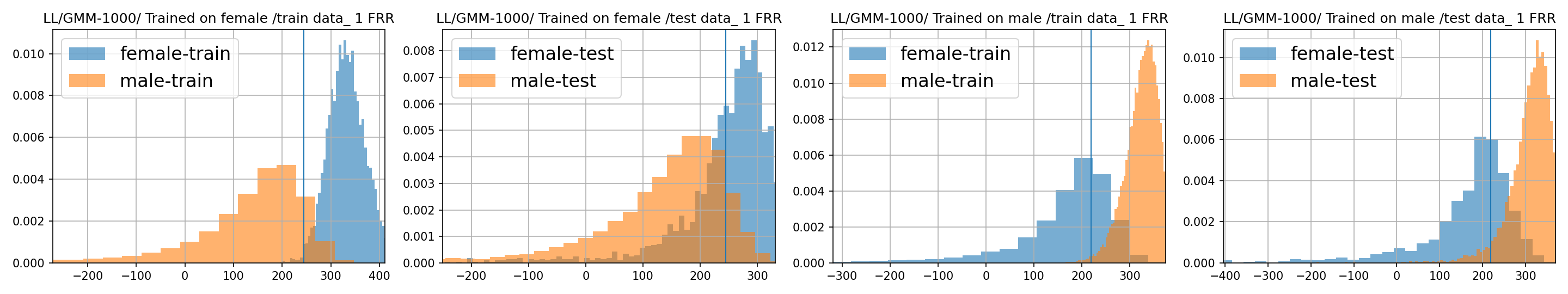}
       \caption{GMMs trained on AEs latent space that itself trained with contrastive loss}
   \end{subfigure}}\\

    \caption{Log-likelihood plots of \emph{1000} component GMMs for various latent spaces and configurations.}
    \label{fig:gmm_plots}
    \vspace{-14pt}
    
\end{figure*}

\subsubsection{Likelihood Visualization}
% In Figure .\ref{fig:gmm_plots} the disentanglement of the different demographic groups in the latentspace is shown, for simplicity of demonstration and comparison purposes we limit ourselves here to the gender demographic i.e. \emph{male} and \emph{female}. Each line in the image corresponds to different latentspaces, The first line corresponds to directly modeling the $\mathcal{W}^{+}$ space of StyleGANv3, we can observe high overlap (high false match rate). The second line, corresponds to modeling the bottleneck layer of an autoencoder without contrastive loss being applied to it. We observe the same high overlap as directly modeling the $\mathcal{W}^{+}$. We can observe our methods' effectiveness by lower the overlap of the gender likelihood plots from the last line.
In \autoref{fig:gmm_plots}, we demonstrate the modeling of different demographic groups in the latent space using likelihood plots. For the sake of simplicity and comparison, we limit this experiment to the gender demographic, which includes \emph{male} and \emph{female}. The first row represents log-likelihood plots for the original $\mathcal{W}^{+}$ space of StyleGAN. The first column corresponds to the LL of the model trained on train subset of the $male$ demographic in the MORPH dataset and the LL of it in comparison to the $female$ demographic of the train subset of the MORPH dataset. The second column is the same experiment except that the GMM is trained on the $female$ demographic and the LL showned in comparison with $male$ demographic. The third and fourth columns are LL plots for models trained on the previous $male$ and $female$ demographics using the train subset and the LL plots are drawn for the test subset. The second row is the same experiment settings as before beside we used the bottleneck output of our AE as modeling space. We can observe our method is effective because the overlap between two distributions ($female$ and $male$) in test cases are significantly reduced.
%the overlaps are significantly reduced when our method is employed.
\vspace{-8pt}
\subsubsection{Implementation Details} 
We used PyTorch for our autoencoder implementation. For the trained StyleGANv3 generator and inversion based on the \cite{psp} we used the model provided by \cite{sg3models_thirdtime} paper. For the GMMs, we used scikit-learn \cite{scikit-learn}. Autoencoder was trained on a single NVIDIA RTX 3090Ti. We optimized our implementation to increase the training batch size as much as possible to minimize the effect caused by the unbalanced appearance of labels in contrastive loss. We did not change the sampling procedure to make the under-represented classes appear more frequently. We set the contribution of each demographic equally (i.e. $c_g = 1$ in \autoref{eq:total_contrastive}). We experimented with different values for $\lambda_1$ and $\lambda_2$ in \autoref{eq:total_loss_autoencoder} and found that setting them to $100$ and $1$, respectively, worked well for a batch size of $192$. We set the number of mixture components, $M$, to $1000$. We determined this through qualitative evaluation of the reconstruction quality (e.g. using grids like in \autoref{fig:autoencoder_disentg_recons}) as well as the contrastive loss employed in the latent space (as depicted in \ref{fig:gmm_plots}). We experimented with two versions of the autoencoder architecture: one using tensor-based encoding and decoding, and the other using a flattened version. We observed that the flattened version performed slightly better. For the encoder part, we used linear layers with dimensions of $8192-4096-2048-1024-512$, with LeakyReLU activations and an initial learning rate of $0.001$.
For the decoder part of the autoencoder, we employed $512-1024-2048-4096-8192$ Linear Layers with LeakyReLU activation functions for all of the layers besides the last one to preserve the range of the input-output of the autoencoder. 
%\href{Hidden}.
% TODO 
%\href{https://gitlab.idiap.ch/safer/aes_template}{Gitlab repostiroy}.
%\href{Hidden}{This} repo.
 
%%%%%%% acknowledgment
\section{Conclusion}
In this work, we present a simple yet effective method for modeling the latent-space of any StyleGAN-based generator. In contrast to previous works that are using much more complex modeling schemes we used simple modeling technique. Our method can be employed to model and later on generate synthetic images according to arbitrary demogrphic groups. One can categorize our proposed method as pre-processing method for addressing bias in existing models.
% We believe the method can be improved significantly by changing the distance function employed in the reconstruction and contrastive loss by Riemannian distance which is more native to the nature of generative models as previously explored in the literature \cite{latentoddity}. 
% We also plan to investigate further the modeling stage with more advanced techniques, such as Normalizing flows.
\section*{Acknowledgment}
This research is based upon work conducted in the project SAFER and supported by the Hasler Foundation under the Responsible AI program.

\clearpage
\clearpage 

{\small
\bibliography{main.bib}
\bibliographystyle{ieee_fullname}
}

\end{document}